\documentclass[11pt]{article}

\usepackage[final]{acl}

\usepackage{times}
\usepackage{latexsym}
\usepackage{graphicx}
\usepackage{booktabs}
\usepackage{multirow}
\usepackage{array}
\usepackage{tikz}
\usepackage{makecell}
\usepackage{subcaption}
\usepackage{hyperref}
\usepackage{placeins}
\usepackage{pgfplots}
\usepackage{tabularx}
\usepackage[ruled,vlined]{algorithm2e}

\pgfplotsset{compat=1.18}
\usepackage[T1]{fontenc}
\usepackage{kotex}
\usepackage[utf8]{inputenc}

\usepackage{microtype}

\usepackage{inconsolata}

\usepackage{graphicx}
\usepackage{multicol}
\usepackage{pifont}
\usepackage{adjustbox}
\usepackage{color}
\usepackage{xcolor}
\usepackage[breakable]{tcolorbox}

\usepackage{caption}

\usepackage[table]{xcolor}
\definecolor{softgray}{gray}{0.93} 
\definecolor{tealframe}{RGB}{0,128,128}
\definecolor{tealtitle}{RGB}{180,226,226}
\definecolor{speechfollow}{RGB}{255,205,178}
\definecolor{speechignore}{RGB}{229,152,155}
\definecolor{speechmis}{RGB}{181,130,140}
\definecolor{steelblue}{RGB}{70,130,180}

       \usepackage{CJKutf8}
       \usepackage{amsmath}
       \usepackage{amssymb}
       
       \newcommand{\cmark}{\textcolor{green!50!black}{$\ding{51}$}}
       \newcommand{\xmark}{\textcolor{red!60!black}{$\times$}}
\title{KoALa-Bench: Evaluating Large Audio Language Models on Korean Speech Understanding and Faithfulness}
\author{
  Jinyoung Kim$^{1*}$, Hyeongsoo Lim$^{1*}$, Eunseo Seo$^{1*}$, Minho Jang$^{1*}$, \\
  \textbf{Keunwoo Choi$^{2}$, Seungyoun Shin$^{2}$, Ji Won Yoon$^{1\dagger}$} \\
  $^{1}$Department of Artificial Intelligence, Chung-Ang University, $^{2}$Upstage AI \\
  \footnotesize\texttt{\{wlsdud338, andrew1001, jeo0534, sunbi8534, jiwonyoon\}@cau.ac.kr} \\
  \footnotesize\texttt{\{keunwoo, logan\}@upstage.ai}
}

\begin{document}
\begin{CJK}{UTF8}{mj}
\maketitle

\renewcommand{\thefootnote}{}
\footnotetext{\textsuperscript{*}Equal contribution}
\footnotetext{\textsuperscript{\dag}Corresponding author}
\renewcommand{\thefootnote}{\arabic{footnote}}
\begin{abstract}
Recent advances in large audio language models (LALMs) have enabled multilingual speech understanding. However, benchmarks for evaluating LALMs remain scarce for non-English languages, with Korean being one such underexplored case. In this paper, we introduce \textbf{KoALa-Bench}, a comprehensive benchmark for evaluating Korean speech understanding and speech faithfulness of LALMs. In particular, KoALa-Bench comprises six tasks. Four tasks evaluate fundamental speech understanding capabilities, including automatic speech recognition, speech translation, speech question answering, and speech instruction following, while the remaining two tasks evaluate speech faithfulness, motivated by our observation that several LALMs often fail to fully leverage the speech modality. Furthermore, to reflect Korea-specific knowledge, our benchmark incorporates listening questions from the Korean college scholastic ability test as well as content covering Korean cultural domains.
%We assess faithfulness from two perspectives, each evaluated through a dedicated task. The first examines whether LALMs ignore the speech input and generate responses based solely on the text input, and the second evaluates whether LALMs correctly utilize information at different positions within long-form speech inputs. 
%추가적으로, 
We conduct extensive experiments across six models, including both white-box and black-box ones. Our benchmark, evaluation code, and leaderboard are publicly available at \url{https://github.com/scai-research/KoALa-Bench}.

%Large audio language models (LALMs)는 큰 발전을 이룸, 또한 다양한 언어를 지원하기 시작했는데, Non-english 능력을 평가할 벤치마크가 없음. 특히 Korean에서는 평가 벤치마크와 평가 지표가 standardized되어있지 않다. 우리는 이를 해결하기 위해 코알라 벤치를 제안하며,  이는 6가지 태스크로 이루어져있으며, 4가지의 기본적인 음성 이해를 위한 태스크와, speech grounding 능력을 평가하기 위한 추가적인 2가지의 task로 이루어짐. while 4가지는 asr, translation, 등으로 이루어져있으며, 

%abstract 1

% 1문단: lalm benchmark 부족 -> 그래서 introduce
% 2문단: 우리의 목적 1) english benchmark를 korean으로 잘 align시키기 2) 기존에 없는 korean speech faithfulness evaluation task 제안 및 evaluation db 만들었음
% 	기존에 있는 english 및 korean을 활용해서 asr, st, sqa, sif 수행 (llm judge 얘기?) / 두번째 이유: lalm이 text bias있음을 발견, korean에서 특히 faithfulness 취약 (근거는 있는지?)
% 3문단: 첫번째 speech faithfulness에 있어서는 ~~ 제안 + 한국 문화
% 4문단: 두번째 faithfulness에서는 position-aware qa 제안 + metric
% 5문단: 실험 많이 했다 (모델명들 ...)

\end{abstract}

\section{Introduction}
% \begin{figure}[t]
%     \centering
%     \includegraphics[width=0.9\columnwidth]{images/teaser.png}
%     \caption{Overview of model performance across six tasks in KoALa-Bench. Each axis represents a task, and the shaded area indicates the normalized score of each model.}
%     \label{fig:teaser}
% \end{figure}
Recent advances in multimodal large language models (MLLMs) \citep{qwen2-audio,tang2024salmonn,audiopalm,team2023gemini} have enabled models to process diverse modalities, including images, videos, and audio. Among these, speech is particularly important as it represents one of the most natural forms of human interaction. Accordingly, large audio language models (LALMs) \citep{navercloudhyperclovaxteam2026hyperclovax8bomni,xu2025qwen3omnitechnicalreport,fang-etal-2025-llamaomni2,qwen2-audio}, which integrate speech encoders with pretrained language models, have emerged as a promising paradigm for speech understanding and spoken interaction, increasingly supporting multilingual speech inputs.

As LALMs expand their multilingual capabilities, comprehensive evaluation frameworks for speech understanding across diverse languages become increasingly important. However, existing benchmarks remain predominantly English-centric, leaving languages such as Korean largely underexplored \citep{wang-etal-2025-audiobench, adu-bench, air-bench}. Meanwhile, existing Korean speech benchmarks are primarily designed for traditional speech processing tasks rather than evaluating the speech understanding capabilities of LALMs \citep{ksponspeech, clovacall, zeroth_korean}. As a result, there is currently a lack of standardized benchmarks specifically designed to evaluate LALMs using Korean speech inputs. To address this limitation, we introduce \textbf{Ko}rean \textbf{A}udio \textbf{La}nguage benchmark, called \textbf{KoALa-Bench}, a comprehensive benchmark for evaluating the Korean speech understanding capabilities of LALMs.

KoALa-Bench consists of six tasks, four for standard speech understanding and two newly proposed for evaluating speech faithfulness.
%For these tasks, we leverage existing Korean datasets and translated English datasets, synthesized into speech, with LLM-based verification to ensure sample quality. 
The four standard tasks cover automatic speech recognition (ASR), speech translation (ST), speech question answering (SQA), and speech instruction following (SIF). For these tasks, we leverage existing Korean datasets as well as English datasets. We align the English datasets with Korean using translation and TTS synthesis, filtering out low-quality samples based on character error rate (CER) or human verification to ensure sample quality.
%Furthermore, 
In addition, we incorporate listening questions from the korean college scholastic ability test (KCSAT) as authentic long-form Korean speech samples, which are included in the SQA task. 
%These tasks consist of speech-aware context question answering (SCA-QA) and position-aware question answering (PA-QA).

The two proposed speech faithfulness tasks, speech-aware context question answering (SCA-QA) and position-aware question answering (PA-QA), are motivated by our observation that LALMs often fail to fully leverage speech input when generating responses. Specifically, we evaluate speech faithfulness from two perspectives. First, SCA-QA examines modality faithfulness by examining whether LALMs fully leverage the speech input or rely solely on the text input. By comparing model responses to a text question with and without a counterfactual speech context, we assess how faithfully LALMs utilize the given speech input. To further reflect Korea-specific knowledge, SCA-QA incorporates crawled Korean cultural domain data covering history, sports, and K-pop. Second, PA-QA assesses positional faithfulness with respect to the position of evidence within long-form speech inputs. To construct the dataset, we identify the location of each answer span in the speech, and measure accuracy at each position. This enables a fine-grained analysis of how faithfully LALMs comprehend information across speech inputs.
%We also incorporate crawled Korean cultural domain data covering history, sports, and K-pop in SCA-QA, enabling evaluation of Korea-specific knowledge.

We conduct extensive experiments across both white-box and black-box models, including Qwen3-Omni \citep{xu2025qwen3omnitechnicalreport}, Gemma-3n \citep{gemma_3n_2025}, GPT-audio \citep{hurst2024gpt-4o}, Gemini-flash \citep{comanici2025gemini-2.5-flash}, and Raon-speech \citep{raon} with varying parameter scales, providing a comprehensive analysis of current LALMs on Korean speech understanding. Our main contributions are summarized as follows:
\begin{itemize}
    \item We introduce KoALa-Bench, a comprehensive benchmark for evaluating the Korean speech understanding capabilities of LALMs. \textit{To the best of our knowledge, this is the first universal benchmark dedicated to evaluating Korean speech understanding and faithfulness of LALMs.}
    \item We propose two novel tasks, SCA-QA and PA-QA, to evaluate the speech faithfulness of LALMs in terms of modality and position, examining whether models fully leverage the given speech input.
    \item To evaluate Korea-specific knowledge, we construct evaluation datasets from KCSAT and crawled cultural domain content covering K-history, K-sports, and K-pop.
\end{itemize}

\begin{figure*}[t!]
\centering

%-------------------- top (a) --------------------%
\begin{subfigure}[t]{\linewidth}
    \centering
    \includegraphics[width=\linewidth]{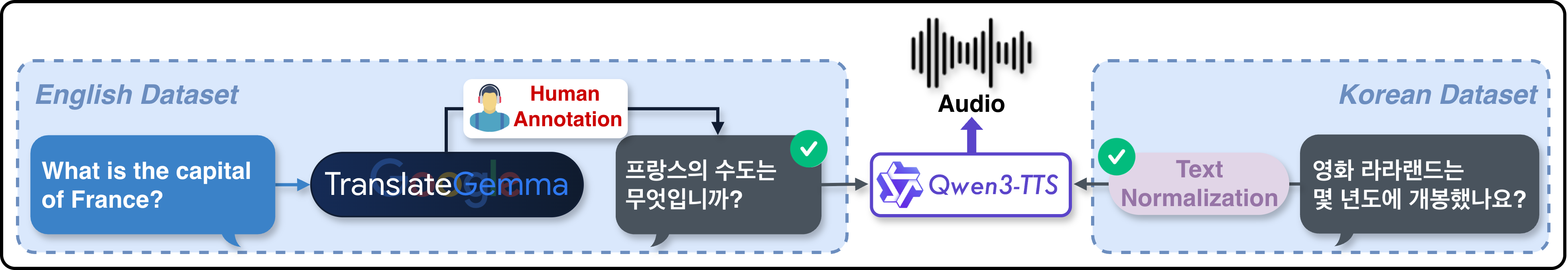}
    \caption{High-level overview of the dataset construction pipeline for standard speech understanding tasks in KoALa-Bench.}
    % \caption{Overview of the fundamental pipeline. To generate audio for text-only datasets, speech synthesis is utilized. Notably, if the dataset consists only of English text, it is translated into Korean using Gemma and then synthesized into audio.} 
\end{subfigure}

\vspace{6pt} 

%------------------ bottom row (b & c) ----------------%
\begin{subfigure}[t]{0.54\linewidth}
    \centering
    \includegraphics[width=\linewidth]{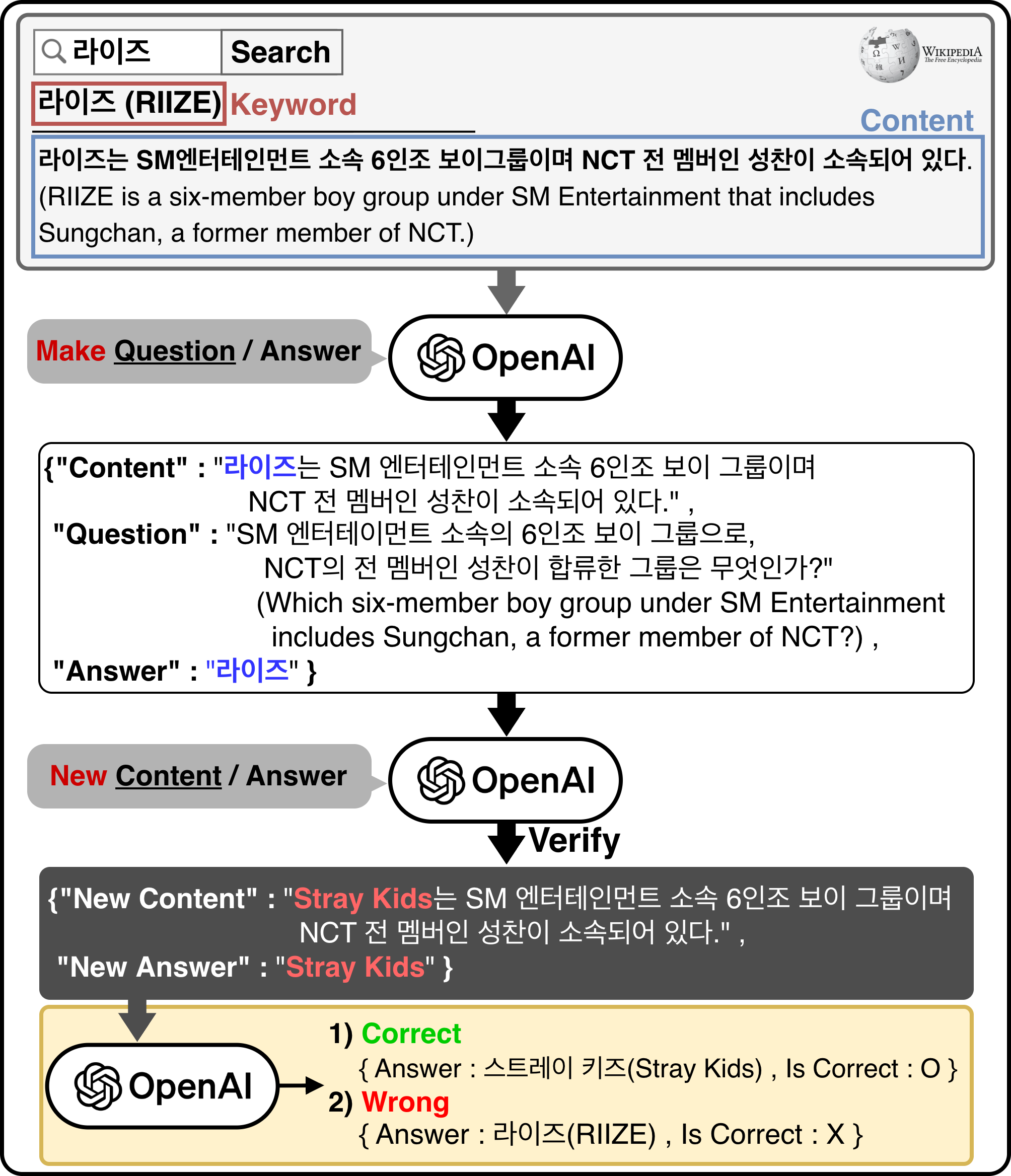}
    \caption{Construction process of the SCA-QA dataset.}
    % \caption{Overview of the SCA-QA construction process. To evaluate whether models rely on the given acoustic context or internal parametric knowledge, we generate paired questions for each context, where one is aligned with common world knowledge and the other intentionally conflicts with it by replacing key entities.} 
\end{subfigure}
\hfill
\begin{subfigure}[t]{0.45\linewidth}
    \centering
    \includegraphics[width=\linewidth]{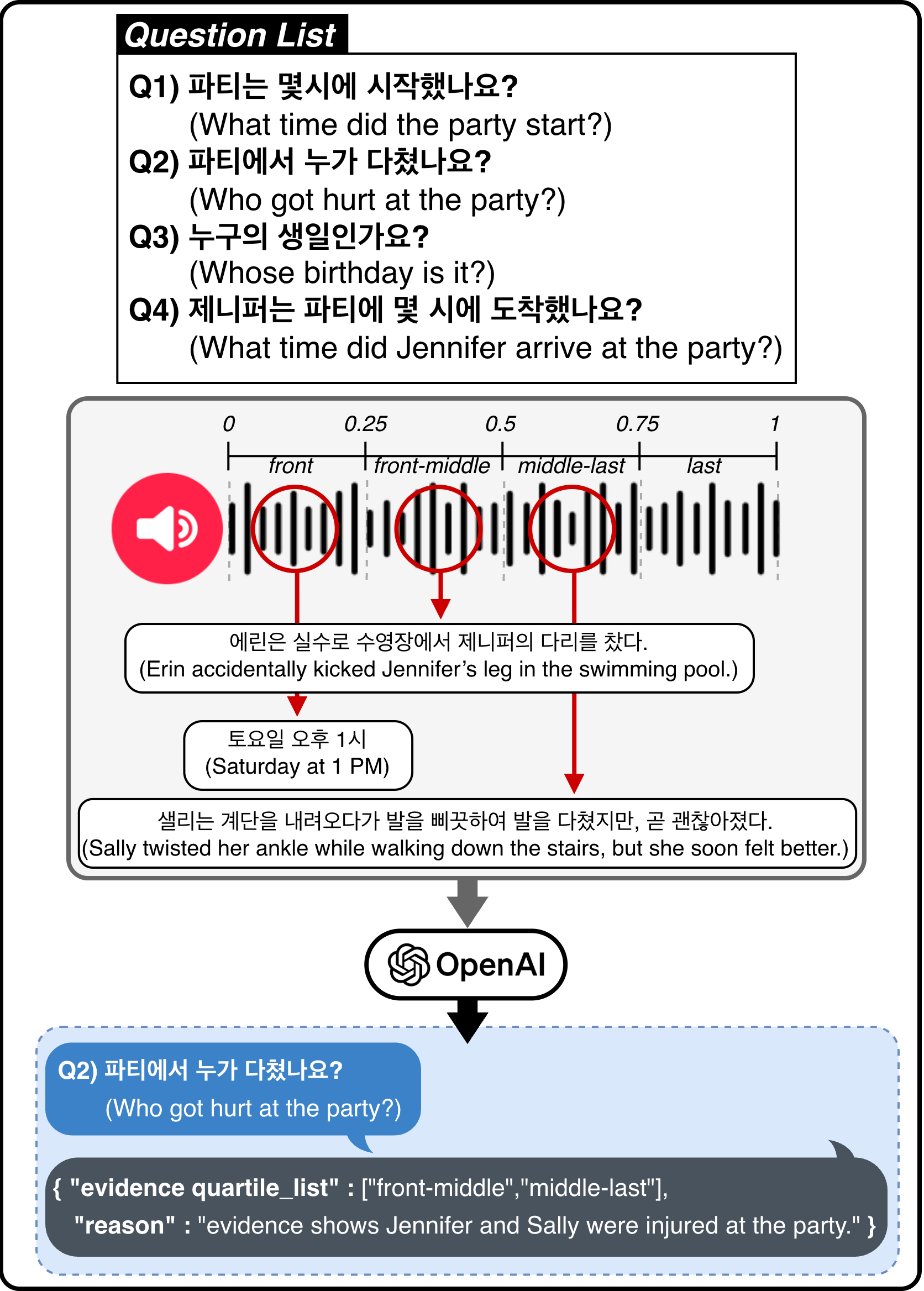}
    \caption{The evidence-annotated dataset generation pipeline for PA-QA.}
\end{subfigure}

\caption{Overview of the KoALa-Bench pipeline. (a) The fundamental pipeline. Synthesizes audio for text-only datasets, translating English text into Korean via Gemma beforehand. (b) The SCA-QA construction process. Generates paired questions (factually aligned vs. intentionally conflicting) to test whether models rely on acoustic context rather than parametric knowledge. (c) The PA-QA dataset generation pipeline. Maps supporting evidence to one of four temporal segments to enable position-aware performance analysis.}
\label{fig:KoALa-Bench pipeline}
\end{figure*}

\section{Related Works}
\paragraph{Audio Benchmarks.}
As LALMs expand their audio understanding capabilities, various benchmarks have been proposed to evaluate their performance on audio-conditioned tasks.
% AIR-Bench \citep{air-bench} evaluates audio understanding and instruction following across diverse audio inputs, while AudioBench \citep{wang-etal-2025-audiobench} provides a comprehensive evaluation suite covering multiple audio tasks. 
AIR-Bench \citep{air-bench} and AudioBench \citep{wang-etal-2025-audiobench} evaluate audio understanding and instruction following across diverse audio inputs. Dynamic-SUPERB \citep{huang2024dynamic} focuses on diverse speech processing tasks including content, speaker, and semantic understanding, while AU-Harness \citep{surapaneni2025harness} focuses on temporal audio understanding and complex reasoning capabilities.
More recent benchmarks explore specialized abilities, including audio dialogue understanding and expert-level reasoning \citep{adu-bench,MMAU}.
However, existing benchmarks remain largely English-centric, limiting their ability to evaluate the speech understanding capabilities of multilingual LALMs, particularly for languages such as Korean. To address this gap, we propose a dedicated benchmark for evaluating Korean speech understanding capabilities of LALMs.

\paragraph{Faithfulness of Multimodal Language Models.}
Recent MLLMs \citep{longcatflashomnitechnicalreport,liu2025voxtral,tang2024salmonn,audiopalm} have demonstrated strong capabilities in integrating information from multiple modalities such as text, images, and audio. 
However, prior studies report that these models often generate responses that are inconsistent with the provided inputs, relying instead on parametric knowledge stored in the language model \cite{bai2024hallucination,faith1_1,faith1_2}. This phenomenon, commonly referred to as multimodal hallucination, reflects a lack of faithfulness to the given modality. 
Faithfulness also becomes challenging when models process long speech inputs, as prior studies report that multimodal models often struggle with long-context audio understanding \citep{faith2_1,faith2_2}.
Despite these observations, the evaluation of faithfulness to speech inputs remains largely underexplored.
To facilitate a systematic analysis of this capability, we introduce two complementary evaluation tasks, SCA-QA and PA-QA, designed to assess how faithfully LALMs utilize the given speech input.

% \paragraph{Faithfulness of Multimodal Language Models.}
% Recent multimodal language models (MLLMs) \citep{qwen2-audio,team2023gemini,tang2024salmonn, rubenstein2023audiopalm} have demonstrated strong capabilities in tasks that integrate information from different modalities, including images, audio and text. However, prior studies \citep{liu-etal-2025-insight, bai2024hallucination, zhu2024unraveling} have shown that multimodal models often rely on parametric knowledge stored in language models while neglecting other modalities, such as images or audio. This limitation leads models to generate incorrect responses while ignoring information from other modalities, highlighting the importance of evaluating whether MLLMs properly utilize multimodal inputs.

\section{KoALa-Bench}
% \paragraph{Overview of KoALa-Bench.}
% ========= 첫 문장 후보
% 후보1: We introduce KoALa-Bench, a benchmark designed to assess the speech understanding, robustness, and speech faithfulness of Korean LALMs.
% 후보 2: We introduce KoALa-Bench, a dedicated benchmark for Korean LALMs.
We introduce KoALa-Bench, a benchmark designed to assess the speech understanding and speech faithfulness of Korean LALMs across six tasks, including ASR, ST, SQA, SIF, SCA-QA, and PA-QA. Following recent LALM benchmarks \citep{chen2026voicebench,wang-etal-2025-audiobench,yan2025uro,adu-bench}, KoALa-Bench includes synthesized speech in several subsets. To ensure data integrity, we conduct quality verification by first removing samples with a CER exceeding 0.1, and then performing human annotation to further filter out unnatural samples. The instructions for human annotation are provided in Appendix \ref{app:human_annotation}. An overview of the data construction pipeline is illustrated in Figure \ref{fig:KoALa-Bench pipeline}, and the detailed statistics of each subset are summarized in Table \ref{tab:benchmark_dataset}. The prompts for all tasks are provided in Appendix \ref{sec:prompts_list}. Below, we describe each subset along with its corresponding evaluation methodology.

% Existing evaluation frameworks primarily focus on generic speech understanding and speech-based QA, with limited consideration of Korean cultural and linguistic characteristics. In addition, prior benchmarks lack evaluation protocols for long duration speech robustness and speech context faithfulness, where speech context faithfulness evaluates whether model predictions are grounded in the acoustic context of the input speech or derived from parametric knowledge.
% To address these limitations, KoALa-Bench evaluates Korean LALMs across six task dimensions: ASR, ST (English-to-Korean), SQA, SIF, SCA-QA, which measures whether models utilize both speech and text information and PA-QA for assessing long-duration speech robustness.
% Furthermore, to evaluate model robustness under noisy conditions, we additionally construct noise augmented versions of several datasets by injecting noise into the speech inputs.

\subsection{Automatic Speech Recognition (ASR)}
The ASR task is the most fundamental speech understanding task, evaluating a model's ability to transcribe spoken utterances into text. The model receives a target speech input along with a text prompt instructing it to transcribe the speech into Korean. The generated transcription is then compared with the ground-truth reference. 

\paragraph{Datasets.}
We use three datasets for ASR evaluation: KsponSpeech \citep{ksponspeech}, Common Voice Korean \citep{commonvoice}, and Zeroth-Korean\footnote{https://openslr.org/40/}. KsponSpeech provides spontaneous Korean conversational speech covering diverse everyday topics. Common Voice Korean and Zeroth-Korean contain speech recorded under diverse acoustic conditions as well as relatively clean read speech, enabling evaluation across different speaking styles and recording environments.

\paragraph{Evaluation.}
ASR performance is evaluated using character error rate (CER), which measures the edit distance between predicted transcriptions and ground-truth references at the character level.
Before computing CER, we apply a normalization pipeline that removes punctuation and collapses whitespace.

\subsection{Speech Translation (ST)}
% Korean LALMs are expected to process both English and Korean. Accordingly, KoALa-Bench includes a cross-lingual speech translation task that translates English speech into Korean text. The task evaluates whether the generated output preserves the semantic content of the source speech while producing fluent Korean text. Evaluation is conducted using automatic similarity measures that assess both lexical overlap and semantic consistency between the generated output and the reference translation.

The ST task evaluates cross-lingual speech translation from English speech into Korean text. We select English as the source language as it is the most frequently encountered foreign language for Korean users, making this the most practically relevant translation direction. The model receives an English speech input along with a text prompt instructing it to translate the speech into Korean.

\paragraph{Datasets.} 
We use the ETRI English–Korean Speech Translation Corpus\footnote{https://epretx.etri.re.kr}, built from TED Talks following the MuST-C\footnote{https://ict.fbk.eu/must-c/} data construction methodology. It provides English speech segments paired with Korean translations for evaluation.
\paragraph{Evaluation.}
We evaluate translation quality using BLEU \citep{bleu}, METEOR \citep{meteor}, and BERTScore \citep{Zhang*2020BERTScore:}. BLEU and METEOR measure n-gram overlap between the generated translation and the reference, while BERTScore evaluates semantic similarity using contextual embeddings. For BERTScore, we use \texttt{xlm-roberta-large} \citep{conneau2020unsupervised} as the backbone model.

\subsection{Speech Question Answering (SQA)}
% The speech understanding task evaluates the model’s ability to answer questions based on acoustic input. In this setting, speech segments are provided as context, which may consist of either natural recordings or speech synthesized using the Qwen3-TTS model \citep{hu2026qwen3ttstechnicalreport}. The model is required to generate responses that correctly follow the textual instruction based on the given speech content. This task evaluates the model’s ability to execute text-based instructions conditioned on the provided acoustic context.
The SQA task evaluates the model's ability to comprehend speech content and answer questions about it. The model receives a speech input as context, paired with a textual question and answer choices, and is required to select the correct option.

\paragraph{Datasets.} For the SQA task, we evaluate on both short-form and long-form speech datasets. For short-form evaluation, we use two Korean QA benchmarks: CLIcK \citep{kim2024click} and KoBEST-BoolQ \citep{kim2022kobest}. CLIcK is a multiple-choice QA benchmark sourced from official exams and textbooks, covering 11 subcategories across language and culture domains. KoBEST-BoolQ is the Boolean QA subset of KoBEST, where each instance consists of a Korean Wikipedia passage and a yes/no question. Since both benchmarks are originally text-based, we synthesize the speech inputs using Qwen3-TTS \citep{hu2026qwen3ttstechnicalreport} with additional quality verification. %(Section \ref{sec:quality_control}).

For long-form evaluation, we construct long speech question and answering (LSQA) from the listening section of the KCSAT, a nationwide university entrance examination known for its high difficulty. We collect the original audio recordings from EBSi\footnote{https://www.ebsi.co.kr/ebs/pot/poti/main.ebs}, which consist of speech passages paired with multiple-choice questions. As the original recordings contain multiple questions in a single long audio file, we segment the audio into question-level samples to construct individual instances. A sample instance from the KCSAT dataset is provided in Appendix~\ref{sec:kcsat_example}.

\paragraph{Evaluation.}
For all SQA datasets, we use accuracy as the evaluation metric, measuring the proportion of correctly answered questions. We adopt two scoring approaches to determine the model's prediction: logit-based and generation-based. The logit-based approach computes the log-probability of each answer choice and treats the one with the highest value as the model's prediction. The generation-based approach extracts the predicted choice from the generated response via string matching.

% \subsubsection{Long Speech Question Answering (LSQA)}
% %K-SAT 내용
% \paragraph{Datasets.}
% To evaluate the ability of LALMs to understand long-form speech, we construct LSQA based on the Korean college scholastic ability test (KCSAT) listening section. The dataset is collected from EBSi\footnote{https://www.ebsi.co.kr/ebs/pot/poti/main.ebs}, an official educational platform providing KCSAT preparation materials.
% LSQA consists of speech passages paired with multiple-choice questions in text form. We segment the original long-form audio recordings to construct individual samples while preserving the long-context nature of the speech inputs.
% \paragraph{Metrics.}
% Performance on LSQA is evaluated using accuracy. Since each question is presented in a multiple-choice format, the model selects one answer from the given options, and accuracy is computed as the proportion of correctly predicted answers.

\subsection{Speech Instruction Following (SIF)}
The SIF task evaluates the model's ability to understand and follow instructions delivered entirely in speech form. Unlike SQA, the model receives only a speech input containing the instruction, without any accompanying text, and is required to generate an appropriate response.

\paragraph{Datasets.}
% In this task, the question is given as audio, and the model is required to generate the answer in text.
% Korean instruction-following test samples are extremely limited. 
% We translated the existing English datasets (ALPACA \cite{alpaca}, Vicuna \cite{vicuna2023}, and OpenHermes\footnote{https://huggingface.co/datasets/teknium/openhermes}) into Korean and synthesized the instructions into audio.
% We also synthesize instructions from KUDGE \cite{kudge}, a Korean instruction-following dataset.
We translate existing English instruction datasets into Korean using TranslateGemma-27B-IT \citep{translateGemma}. The source datasets include ALPACA \citep{alpaca}, Vicuna \citep{vicuna2023}, and OpenHermes\footnote{https://huggingface.co/datasets/teknium/openhermes}. We additionally include Korean-native instructions from KUDGE \citep{kudge}. All instructions are then synthesized into speech using Qwen3-TTS \citep{hu2026qwen3ttstechnicalreport} with additional quality verification.
\paragraph{Evaluation.}
% Instruction-following performance is evaluated using GPT-4o as a judge model. The judge assesses whether the generated response correctly follows the instruction and produces an appropriate output.
SIF performance is evaluated using GPT-4o as a judge model, following prior work on instruction following evaluation \citep{kim2025biggen,kim2024prometheus,wang-etal-2025-audiobench}. The judge model assesses whether the generated response correctly follows the instruction. Detailed evaluation prompts are provided in Appendix~\ref{sec:llm_as_judge}.

\subsection{Speech Context-Aware Question Answering (SCA-QA)}
% 원본: Motivated by DisentQA \citep{neeman-etal-2023-disentqa}, we introduce SCA-QA, which constructs paired questions per speech context to determine whether model responses rely on acoustic context or parametric knowledge.
Motivated by our observation, as illustrated in Appendix~\ref{app:failure_speech_context}, that models often ignore speech context, we introduce SCA-QA. Inspired by DisentQA \citep{disentqa}, the task embeds conflicting facts into the speech context and poses questions about them, evaluating whether the model grounds its response in the given acoustic context or relies on parametric knowledge.
\paragraph{Datasets.}
To construct the dataset, we generate 200 question-answer pairs for each topic using the GPT API, covering K-pop, K-history, and K-sports. We then replace key answer entities in both the question and the speech context with different entities (e.g., replacing “BTS'' with “BLACKPINK''), so that the correct answer based on the speech context conflicts with world knowledge. This allows us to evaluate whether the model faithfully grounds its answer in the given speech context rather than relying on its parametric knowledge. The construction pipeline and prompts are provided in Appendix~\ref{sec:sca_qa_prompts}. All samples were filtered through human annotation, where annotators assessed the appropriateness of each QA pair and the validity of the conflicting facts. The model receives a speech context containing the conflicting fact, along with a textual question and four answer choices. A detailed example is provided in Appendix~\ref{sec:sca_qa_example}.
\paragraph{Evaluation.}
We evaluate model responses using speech context faithfulness (SCF), which measures whether predictions rely on the provided speech context rather than parametric knowledge. For each sample, we first identify cases where the model answers the textual question correctly, and then evaluate the same question with the corresponding speech context. If the model changes its answer according to the speech content when the contextual information conflicts, the response is considered context-faithful. The final score is computed as the proportion of such context-faithful samples. We provide the formal definition in Appendix~\ref{sec:evaluation_appendix}.

% ============== koala benchmark task table

\begin{table*}[t]
\centering
\footnotesize
\setlength{\tabcolsep}{6pt}
\renewcommand{\arraystretch}{1.0}

\begin{adjustbox}{max width=\textwidth}
\begin{tabular}{l|l| c| c|c c c c c}
\toprule\hline
\multicolumn{2}{c|}{\multirow{2}{*}{\textbf{Dataset}}} &
\multirow{2}{*}{\textbf{Metrics}} &
\multirow{2}{*}{\textbf{Noise}} &
\multirow{2}{*}{\textbf{\# Samples}} &
\multirow{2}{*}{\textbf{Hours(h)}} &
\multicolumn{3}{c}{\textbf{Length(s)}} \\
\cline{7-9}
\multicolumn{2}{c|}{} & & &  &  & \textbf{Avg} & \textbf{Min} & \textbf{Max} \\

\hline\hline
\multicolumn{9}{c}{\cellcolor{softgray}\textbf{ASR}}\\
\hline\hline
\multicolumn{2}{l|}{Zeroth} & \multirow{4}{*}{CER $\downarrow$} & \ding{51} & 457 & 1.191 & 9.381 & 4.558 & 20.453 \\
\multicolumn{2}{l|}{Common Voice} &  & \ding{51} & 523 & 0.726 & 4.998 & 1.728 & 12.564 \\
\multicolumn{2}{l|}{KsponSpeech-Clean} &  & \ding{55} & 3000 & 2.644 & 3.173 & 0.286 & 19.357 \\
\multicolumn{2}{l|}{KsponSpeech-Other} &  & \ding{55} & 3000 & 3.803 & 4.563 & 0.399 & 20.093 \\

\hline\hline
\multicolumn{9}{c}{\cellcolor{softgray}\textbf{ST}}\\
\hline\hline
\multicolumn{2}{l|}{ETRI-TST-Common} & 
\multirow{2}{*}{\makecell{BLEU / METEOR \\ / BERTScore} $\uparrow$} 
& \ding{55} & 2532 & 4.03 & 5.73 & 	0.32 & 	39.62 \\
\multicolumn{2}{l|}{ETRI-TST-HE} &  & \ding{55} & 544 & 1.07 & 7.09 & 0.77 & 40.79 \\

\hline\hline
\multicolumn{9}{c}{\cellcolor{softgray}\textbf{SQA}}\\
\hline\hline
\multirow{2}{*}{Short-Form} & CLIcK & \multirow{3}{*}{Accuracy $\uparrow$} & \ding{51} & 1315 & 4.180 & 11.444 & 1.600 & 275.402 \\
& KoBest BoolQ &  & \ding{51} & 1404 & 11.950 & 30.642 & 4.640 & 88.308 \\
\cline{1-2}\cline{4-9}
Long-Form & KCSAT &  & \ding{51} & 82 & 2.919 & 128.182 & 62.976 & 208.020 \\

\hline\hline
\multicolumn{9}{c}{\cellcolor{softgray}\textbf{SIF}}\\ 

% \multirow{4}{*}{\makecell[l]{Speech Instruction \\ Following}}

\hline\hline
\multicolumn{2}{l|}{KUDGE} & \multirow{4}{*}{\makecell{Score \\ (GPT as Judge)} $\uparrow$} & \ding{51} & 557 & 1.863 & 10.365 & 3.440 & 23.497 \\
\multicolumn{2}{l|}{Vicuna} &  & \ding{51} & 70 & 0.192 & 8.623 & 2.640 & 16.297 \\
\multicolumn{2}{l|}{OpenHermes} &  & \ding{51} & 78 & 0.174 & 6.450 & 1.680 & 18.617 \\
\multicolumn{2}{l|}{Alpaca} &  & \ding{51} & 69 & 0.111 & 4.375 & 1.760 & 9.017 \\

\hline\hline
\multicolumn{9}{c}{\cellcolor{softgray}\textbf{SCA-QA}}\\
\hline\hline
\multicolumn{2}{l|}{K-history (Before Chosun)} & 
\multirow{4}{*}{SCF Score $\uparrow$} & \ding{55}
& 101 & 0.815 & 29.053 & 11.280 & 61.280 \\
\multicolumn{2}{l|}{K-history (After Chosun)} &  & \ding{55} & 82 & 0.997 & 43.780 & 3.520 & 100.000 \\
\multicolumn{2}{l|}{K-sports} &  & \ding{55} & 88 & 1.698 & 69.488 & 20.400 & 92.640 \\
\multicolumn{2}{l|}{K-pop} &  & \ding{55} & 103 & 1.457 & 50.926 & 8.000 & 82.400 \\
% \multirow{2}{*}{K-History (Before)} & Correct Context & 
% \multirow{8}{*}{\makecell{Speech Context \\ Faithfulness}} & \ding{51}
% & -- & -- & -- & -- & -- \\
%  & Incorrect Context &  & \ding{51}
% & 101 & 0.815 & 29.053 & 11.280 & 61.280 \\
% \cline{1-2}\cline{4-9}
% \multirow{2}{*}{K-History (After Chosun)} & Correct Context &  & \ding{51} & -- & -- & -- & -- & -- \\
%  & Incorrect Context &  & \ding{51} & 82 & 0.997 & 43.780 & 3.520 & 100.000 \\
% \cline{1-2}\cline{4-9}
% \multirow{2}{*}{K-sports} & Correct Context &  & \ding{51} & -- & -- & -- & -- & -- \\
%  & Incorrect Context &  & \ding{51} & 88 & 1.698 & 69.488 & 20.400 & 92.640 \\
% \cline{1-2}\cline{4-9}
% \multirow{2}{*}{K-pop} & Correct Context &  & \ding{51} & -- & -- & -- & -- & -- \\
%  & Incorrect Context &  & \ding{51} & 103 & 1.457 & 50.926 & 8.000 & 82.400 \\
\hline\hline
\multicolumn{9}{c}{\cellcolor{softgray}\textbf{PA-QA}}\\
\hline\hline
\multicolumn{2}{l|}{MCTest} & Accuracy $\uparrow$ & \ding{51} & 325 & 8.732 & 96.142 & 38.400 & 205.120 \\

\hline
\bottomrule
\end{tabular}
\end{adjustbox}

\caption{Datasets used in the proposed KoALa-Bench.}
\label{tab:benchmark_dataset}
\end{table*}
% ============== koala benchmark task table

\subsection{Postion Aware Question Answering (PA-QA) }
% Existing speech benchmarks largely rely on short utterances and lack supporting evidence annotations for each QA pair, making it difficult to analyze which part of the speech context a question depends on. To address this limitation, KoALa-Bench introduces a position-aware evaluation setup by identifying supporting evidence sentences for each question. We construct an evidence annotation pipeline using GPT-4.1-mini to detect supporting sentences and map them to their relative positions within the speech context. Each context is normalized to the range [0,1] and divided into four segments: front (0–0.25), front-middle (0.25–0.5), middle-late (0.5–0.75), and late (0.75–1.0). This dataset enables position-aware analysis of model performance across long speech inputs. The full evidence grounding procedure is described in Appendix \ref{sec:appendix_grounding}.

% Existing speech benchmarks often use short utterances and do not provide evidence annotations for each QA pair, making it difficult to identify which part of the speech input is needed to answer a question. To address this limitation, KoALa-Bench adds a position-aware evaluation datasets. 

Existing speech benchmarks often rely on short utterances and lack evidence-position annotations for each QA pair \citep{chen2026voicebench,wang-etal-2025-audiobench}, making it difficult to analyze which parts of a speech input a model effectively uses or fails to use when answering a question. To address this limitation, KoALa-Bench introduces PA-QA, a position-aware evaluation setup designed to examine model performance according to the location of answer-relevant evidence within long speech inputs.

\paragraph{Datasets.}
We construct the dataset from MCTest \citep{richardson-etal-2013-mctest}, a machine comprehension dataset of short stories and multiple-choice questions. For each QA pair, we first split the speech context into individual sentences, treating them as possible evidence candidates. We then use the question and answer to select the sentences that support the answer, with GPT-4o-mini used for candidate selection and verification. Each verified evidence sentence is mapped to its relative position in the speech context, which is normalized to [0,1] and divided into four temporal segments: front (0--0.25), front-middle (0.25--0.5), middle-last (0.5--0.75), and last (0.75--1.0). This setup allows us to analyze whether model performance changes depending on where the relevant information appears in long speech inputs. The full evidence grounding procedure is described in Appendix \ref{sec:appendix_grounding}. Speech contexts are synthesized from textual passages using Qwen3-TTS \citep{hu2026qwen3ttstechnicalreport}, and supporting evidence sentences are identified with GPT-4o-mini. Also, all samples were verified through the quality control process. A sample instance is provided in Appendix~\ref{sec:pa_qa_example}.

\paragraph{Evaluation.}
We adopt two evaluation methods for PA-QA. The first follows the same protocol  as in SQA and report accuracy over all QA pairs, scoring whether the model selects the correct answer. For position-aware analysis, we group QA pairs by the temporal segment containing their supporting evidence and report segment-wise accuracy, allowing us to examine whether models are more vulnerable to relevant information appearing at specific positions in the speech input.

% \subsection{Quality Control}

% We apply both automatic and human verification to ensure the quality of the synthesized speech and generated question-answer pairs. Following recent LALM benchmarks that use synthesized speech for scalable benchmark construction~\citep{chen2026voicebench,wang-etal-2025-audiobench,yan2025uro,adu-bench}, we construct speech inputs with TTS but validate the outputs before inclusion. For each synthesized utterance, we compute the CER between the input text and the transcription of the synthesized audio, and discard samples whose CER exceeds a predefined threshold. Human annotators further remove samples that sound unnatural or contain audible synthesis artifacts. For SCA-QA, annotators manually review each generated question-answer pair and its entity-replaced counterpart, removing samples with incorrect, inconsistent, or unnatural content. Thus, all TTS-generated speech and LLM-generated QA pairs used in KoALa-Bench undergo automatic or human verification before being included in the benchmark.

\section{Experiments}

\begin{table*}[t]
\centering
\begin{adjustbox}{max width=\textwidth}
\begin{tabular}{l|l|c|@{\hskip 4pt}c@{\hskip 4pt}c@{\hskip 4pt}c@{\hskip 4pt}c@{\hskip 4pt}c@{\hskip 4pt}c}
\toprule\hline
\multicolumn{2}{l|}{\textbf{Dataset}} & \textbf{Metric} & \textbf{Qwen3-omni} & \textbf{Gemma-3n} & \textbf{GPT-audio} & \textbf{Voxtral} & \textbf{Gemini-flash} & \textbf{Raon} \\
\hline\hline
\rowcolor{softgray}
\multicolumn{9}{c}{\textbf{ASR}} \\
\hline\hline
\multicolumn{2}{l|}{Zeroth} & \multirow{4}{*}{CER $\downarrow$} & \textbf{3.39} / \textbf{\textcolor{blue}{4.02}} & 100$\uparrow$ / \textcolor{blue}{100$\uparrow$} & 6.87 / \textcolor{blue}{9.00} & 46.29 / \textcolor{blue}{42.91} & 13.60 / \textcolor{blue}{14.56} & 4.28 / \textcolor{blue}{5.27} \\
\multicolumn{2}{l|}{Common Voice} & & 4.99 / \textcolor{blue}{\textbf{7.02}} & 100$\uparrow$ / \textcolor{blue}{100$\uparrow$} & 33.05 / \textcolor{blue}{36.21} & 65.80 / \textcolor{blue}{63.57} & 13.74 / \textcolor{blue}{26.74} & \textbf{4.78} / \textcolor{blue}{7.39} \\
\multicolumn{2}{l|}{KsponSpeech-Clean} & & 8.58 & 100$\uparrow$  & 100$\uparrow$  & 67.83  & 83.19 & \textbf{8.16} \\
\multicolumn{2}{l|}{KsponSpeech-Other} & & 8.15 & 100$\uparrow$  & 63.64 & 63.15  & 45.14 & \textbf{7.81} \\
\hline\hline
\rowcolor{softgray}
\multicolumn{9}{c}{\textbf{ST}} \\
\hline\hline
\multicolumn{2}{l|}{ETRI-TST-COMMON} & \multirow{2}{*}{BERTScore $\uparrow$} & \textbf{93.38} & 85.02 & 93.10 & 92.64 & 91.60 & 92.67 \\
\multicolumn{2}{l|}{ETRI-TST-HE} & & \textbf{93.89} & 85.82 & 93.69 & 92.98 & 92.17 & 93.22 \\
\hline\hline
\rowcolor{softgray}
\multicolumn{9}{c}{\textbf{SQA}} \\
\hline\hline
\multirow{2}{*}{Short-Form} & CLIcK & \multirow{3}{*}{Accuracy $\uparrow$} & 63.57 / \textcolor{blue}{62.01} & 35.11 / \textcolor{blue}{35.31} & 61.64 / \textcolor{blue}{60.07} & 41.94 / \textcolor{blue}{41.88} & \textbf{67.27} / \textbf{\textcolor{blue}{66.69}} & 55.68 / \textcolor{blue}{55.18} \\
& KoBEST BoolQ & & 50.40 / \textcolor{blue}{49.84} & 50.36 / \textcolor{blue}{50.27} & 51.88 / \textcolor{blue}{50.45} & 50.49 / \textcolor{blue}{50.40} & \textbf{52.86} / \textbf{\textcolor{blue}{54.92}} & 50.45 / \textcolor{blue}{50.51} \\
\cline{1-2}\cline{4-9}
Long-Form & KCSAT & & 80.59 / \textbf{\textcolor{blue}{82.36}} & 33.53 / \textcolor{blue}{37.65} & 52.90 / \textcolor{blue}{47.10} & 69.41 / \textcolor{blue}{71.76} & \textbf{81.18} / \textcolor{blue}{78.82} & 45.88 / \textcolor{blue}{45.29} \\
\hline\hline
\rowcolor{softgray}
\multicolumn{9}{c}{\textbf{SIF}} \\
\hline\hline
\multicolumn{2}{l|}{KUDGE} & \multirow{4}{*}{\makecell{Score \\ (GPT as Judge)} $\uparrow$} & 70.89 / \textcolor{blue}{70.50} & 70.68 / \textcolor{blue}{70.53} & \textbf{74.07} / \textbf{\textcolor{blue}{73.99}} & 59.14 / \textcolor{blue}{59.07} & 70.29 / \textcolor{blue}{70.39} & 68.35 / \textcolor{blue}{66.85} \\
\multicolumn{2}{l|}{Vicuna} & & 76.50 / \textcolor{blue}{71.79} & 79.38 / \textcolor{blue}{78.84} & \textbf{82.14} / \textbf{\textcolor{blue}{81.79}} & 62.82 / \textcolor{blue}{64.86} & 76.43 / \textcolor{blue}{73.29} & 73.81 / \textcolor{blue}{71.52} \\
\multicolumn{2}{l|}{OpenHermes} &  & 84.09 / \textcolor{blue}{80.74} & 82.93 / \textcolor{blue}{84.25} & \textbf{89.42} / \textbf{\textcolor{blue}{89.62}} & 65.82 / \textcolor{blue}{64.87} & 82.69 / \textcolor{blue}{80.71} & 78.65 / \textcolor{blue}{78.30} \\
\multicolumn{2}{l|}{Alpaca} & & 82.55 / \textcolor{blue}{80.68} & 82.83 / \textcolor{blue}{82.20} & \textbf{90.58} / \textbf{\textcolor{blue}{90.58}} & 68.48 / \textcolor{blue}{68.16} & 85.94 / \textcolor{blue}{86.88} & 76.34 / \textcolor{blue}{75.80} \\
\hline
\bottomrule
\end{tabular}
\end{adjustbox}
\caption{Comparison across ASR, ST, SQA, and SIF. All results are averaged over four prompts using greedy decoding. ASR, SQA, and SIF are evaluated on clean and noisy audio. KsponSpeech-Clean/Other denote subsets. $100\uparrow$ indicates CER $>$ 100\%. Blue indicates noisy results.}
\label{tab:all_results}
\end{table*}

\begin{table*}[t!]
\centering
\begin{adjustbox}{max width=\textwidth}
\begin{tabular}{l|l|c|@{\hskip 4pt}c@{\hskip 4pt}c@{\hskip 4pt}c@{\hskip 4pt}c@{\hskip 4pt}c@{\hskip 4pt}c}
\toprule\hline
\multicolumn{2}{l|}{\textbf{Dataset}} & \textbf{Metric} & \textbf{Qwen3-omni} & \textbf{Gemma-3n} & \textbf{GPT-audio} & \textbf{Voxtral} & \textbf{Gemini-flash} & \textbf{Raon} \\
\hline\hline
\multicolumn{9}{c}{\cellcolor{softgray}\textbf{SCA-QA}}\\
\hline\hline
\multicolumn{2}{l|}{K-history (Before)}   & \multirow{4}{*}{\makecell{Text-only Accuracy $\uparrow$}} & 56.44  & 46.78  & 69.30  & 38.37 & \textbf{79.21} & 50.25 \\
\multicolumn{2}{l|}{K-history (After Chosun)}    & & 77.74 & 67.38  & 79.30 & 47.26 & \textbf{87.80} & 61.59 \\
\multicolumn{2}{l|}{K-sports}             & & 52.56 & 55.68 & 69.30  & 30.40  & \textbf{78.41} & 48.58 \\
\multicolumn{2}{l|}{K-pop}                & & 59.47 & 61.17 & 69.90 & 36.89  & \textbf{89.32} & 51.70 \\
\hline
\multicolumn{2}{l|}{K-history (Before)}   &  \multirow{4}{*}{SCF Score $\uparrow$} & \textbf{93.86} & 49.72 & 61.40 & 77.24 & 66.25 & 36.91 \\
\multicolumn{2}{l|}{K-history (After Chosun)}    & & \textbf{85.46} & 36.10 & 32.30 & 76.57 & 59.72 & 21.27 \\
\multicolumn{2}{l|}{K-sports}             & & 84.34 & 69.33 & 39.30 & \textbf{88.85} & 86.96 & 21.89 \\
\multicolumn{2}{l|}{K-pop}                & & \textbf{90.60} & 61.88 & 37.50 & 87.39 & 78.26 & 25.10 \\
\hline\hline
\multicolumn{9}{c}{\cellcolor{softgray}\textbf{PA-QA}}\\
\hline\hline
\multicolumn{2}{l|}{MC-Test Total} & \multirow{5}{*}{Accuracy $\uparrow$} & \textbf{92.51} / \textbf{\textcolor{blue}{93.12}} & 48.70 / \textcolor{blue}{48.78} & 77.23 / \textcolor{blue}{79.69} & 84.25 / \textcolor{blue}{85.24} & 92.31 / \textcolor{blue}{92.00} & 40.98 / \textcolor{blue}{41.28} \\
\cline{1-2}\cline{4-9}
\multirow{4}{*}{\makecell{Position\\- Aware}} & Front &  & 91.55 / \textcolor{blue}{92.07} & 46.72 / \textcolor{blue}{46.55} & 72.05 / \textcolor{blue}{75.78} & 84.31 / \textcolor{blue}{85.34} & \textbf{95.03} / \textbf{\textcolor{blue}{93.17}} & 43.48 / \textcolor{blue}{42.03} \\
& Front-Middle  & & \textbf{93.15} / \textbf{\textcolor{blue}{94.35}} & 50.30 / \textcolor{blue}{50.00} & 83.52 / \textcolor{blue}{83.52} & 88.10 / \textcolor{blue}{88.10} & 89.01 / \textcolor{blue}{87.91} & 42.70 / \textcolor{blue}{44.94} \\
& Middle-Last  & & \textbf{90.26} / \textbf{\textcolor{blue}{89.29}} & 48.38 / \textcolor{blue}{50.00} & 74.39 / \textcolor{blue}{70.73} & 82.79 / \textcolor{blue}{84.42} & 82.93 / \textcolor{blue}{85.37} & 32.47 / \textcolor{blue}{32.47} \\
& Last  & & \textbf{95.31} / \textcolor{blue}{95.83} & 44.79 / \textcolor{blue}{43.75} & 80.39 / \textcolor{blue}{86.27} & 83.85 / \textcolor{blue}{78.65} & 92.16 / \textbf{\textcolor{blue}{98.04}} & 40.00 / \textcolor{blue}{38.18} \\
\hline
\bottomrule
\end{tabular}
\end{adjustbox}
\caption{Performance comparison on speech faithfulness tasks. All results are averaged over four prompts using greedy decoding. SCA-QA reports text-only accuracy and SCF score. PA-QA reports accuracy by evidence position. Blue indicates results on the dataset with additional background noise.}
\label{tab:faithfulness_results}
\end{table*}

\subsection{Evaluation setup}

We evaluated six models, including Qwen3-Omni-30B-A3B-Instruct \citep{xu2025qwen3omnitechnicalreport}, Google-gemma-3n-E4B-it \citep{gemma_3n_2025}, Voxtral-Mini-3B-2507 \citep{liu2025voxtral}, GPT-audio-mini\footnote{https://developers.openai.com/api/docs/models/gpt-audio-mini}, Gemini-flash-lite \citep{comanici2025gemini-2.5-flash} and Raon-Speech-9B \citep{raon}. For open-weight models, inference was performed using vLLM \citep{vllm} on two RTX Pro 6000 Blackwell GPUs with 96 GiB memory each. For proprietary models, we used the official API. The datasets used for evaluation are summarized in Table~\ref{tab:benchmark_dataset}. 

To assess robustness under acoustic degradation, we additionally evaluated models on noise-augmented samples constructed by injecting background noise from the SELECTSTAR noise dataset\footnote{https://open.selectstar.ai/ko/cochl}, which contains diverse environmental sounds such as traffic and ambient location noise. Further details on the noise sources and the augmentation procedure are provided in the Appendix~\ref{sec:noise_dataset_injection}.

% In terms of the evaluation protocol, since the models under evaluation incorporate large language models as their backbone, they can be highly sensitive to prompt phrasing \cite{prompt_sensitivity1, prompt_sensitivity2, prompt_sensitivity3}. To account for this, we evaluate each task using four Korean prompts expressing the same instruction with slight lexical and structural variations, and report the averaged result as the final score. The full prompt list and best results are provided in Appendix~\ref{sec:prompts_list}.
Since LALMs can be highly sensitive to prompt phrasing~\cite{prompt_sensitivity1,prompt_sensitivity2,prompt_sensitivity3}, we evaluate each task using four Korean prompts with slight lexical and structural variations and report the average score as the main result. The full prompt list, including best scores for diagnostic purposes, is provided in Appendix~\ref{sec:prompts_list}.

\subsection{Main Results}
% gpt는 짧은 단어는 못하더라 -> 예를 들어 오디오를 다시 달라고 응답하는 흥미로운 결과를 보았다. 
\subsubsection{Standard Speech Understanding}
Table~\ref{tab:all_results} presents the overall performance of various LALMs across ASR, ST, SQA, and SIF tasks. In ASR, Qwen3 and Raon consistently outperformed other models, with Qwen3 achieving the lowest CER on read speech (Zeroth) and Raon excelling on spontaneous and conversational speech (Common Voice, KsponSpeech). Qwen3 also achieved the highest BERTScore on ST, while Gemini-flash led on short-form SQA benchmarks and Qwen3 on the long-form KCSAT. For SIF, GPT achieved the highest scores across all datasets, indicating stronger instruction-following capabilities. Under noisy conditions, ASR tasks showed noticeable CER increases, whereas SQA and SIF exhibited modest degradation, suggesting that background noise affects low-level recognition more than higher-level understanding.

\subsubsection{Speech Faithfulness}
\paragraph{SCA-QA.}
% As shown in Table~\ref{tab:faithfulness_results}, GPT-audio-mini achieved the highest Text-only accuracy across domains, indicating strong Korea-specific parametric knowledge. However, under counterfactual speech context, it tended to follow its prior knowledge rather than the speech input, suggesting limited use of the speech modality. Qwen3 maintained high Text-only accuracy while exhibiting the lowest prediction preservation, indicating the most effective use of speech input. Voxtral showed relatively stronger speech faithfulness despite lower Text-only accuracy. 
% We further evaluated models using correct speech inputs to verify modality alignment. Additional analysis is provided in the Section \ref{sec:analysis}.
As shown in Table~\ref{tab:faithfulness_results}, SCA-QA highlights a gap between Korea-specific parametric knowledge and speech-grounded faithfulness. Gemini-flash achieved the highest Text-only accuracy across all domains, and GPT-audio also showed strong Text-only performance; however, both obtained much lower SCF scores than Qwen3-Omni and Voxtral, indicating reliance on prior knowledge rather than counterfactual speech evidence. In contrast, Qwen3-Omni achieved the highest SCF score in three of four domains, while Voxtral showed strong faithfulness on K-sports despite lower Text-only accuracy. Raon performed poorly in both Text-only accuracy and SCF, suggesting limited ability to identify the answer from counterfactual speech input. Additional analysis is provided in Section~\ref{sec:analysis}.

\paragraph{PA-QA.}
% On the MC-Test benchmark (Table~\ref{tab:faithfulness_results}), Qwen3-Omni achieved the highest accuracy under both clean and noisy conditions, with particularly stronger performance under noise. Gemma-3n exhibited a monotonic decline in accuracy across the positional analysis, indicating a stronger reliance on early speech information. GPT-audio-mini performed particularly poorly in the Front-Middle region, whereas Voxtral showed weaknesses in the Front-Middle and Late segments.
On PA-QA (Table\ref{tab:faithfulness_results}), we report overall MC-Test and position-aware accuracy to examine evidence grounding across speech positions. Qwen3-Omni achieved the highest overall accuracy under both clean and noisy conditions, with the best performance in Front-Middle and Middle-Last, while Gemini-flash was strongest in Front and noisy Last. These results suggest that Qwen3-Omni provides more consistent grounding across positions, whereas Gemini-flash is particularly effective for early or noisy late evidence. Gemma-3n and Raon remained substantially weaker, indicating limited long-form speech understanding and evidence grounding. Additional analysis is provided in Section~\ref{sec:analysis}.

% === sca-qa 케이스 정리 table
\begin{table}[t]
\centering
\small
\setlength{\tabcolsep}{3pt}
\renewcommand{\arraystretch}{1.2}
\begin{adjustbox}{max width=\columnwidth}
\begin{tabular}{l|l|cccc}
\toprule
\hline
\multicolumn{2}{l|}{} & \textbf{Case 1} & \textbf{Case 2} & \textbf{Case 3} & \textbf{Case 4} \\ 
\hline\hline
\multicolumn{6}{c}{\cellcolor{softgray}\textbf{Speech Context}} \\
\hline\hline
\multicolumn{2}{l|}{Wrong} & O & O & X & X \\
\hline
\multicolumn{2}{l|}{Correct} & O & X & O & X \\
\hline\hline
\multicolumn{6}{c}{\cellcolor{softgray}\textbf{Outcome}} \\
\hline\hline
\multicolumn{2}{l|}{Aligned} & \ding{51} & \ding{55} & \ding{51} & \ding{55} \\
\hline
\multicolumn{2}{l|}{Speech Utilization} & \ding{51} & -- & \ding{51} & -- \\
\hline
\bottomrule
\end{tabular}
\end{adjustbox}
\caption{Case analysis for speech-modality alignment. Case categorization is performed only for instances where the text-context condition is correctly answered.}
\label{tab:alignment_cases}
\end{table}

\section{Analysis}
\label{sec:analysis}
\paragraph{SCA-QA.}
To further assess speech-modality alignment, we compared model predictions under wrong and correct speech contexts. Table~\ref{tab:alignment_cases} summarizes the possible prediction patterns and their interpretations. Case~1 indicated that the model followed the speech input, whereas Case~3 suggested that the speech context was ignored. In contrast, Cases~2 and~4 corresponded to misalignment, where the model failed to produce the correct answer even under the correct speech context. Figure~\ref{fig:alignment_behavior_correct_speech} shows that Voxtral and Raon exhibit speech--text alignment failures, with Raon showing a strong tendency to ignore the speech context. This demonstrates that SCA-QA enables fine-grained evaluation of speech faithfulness, extending to speech--text alignment analysis.

\begin{figure}[t]
\centering
\resizebox{\columnwidth}{!}{
\begin{tikzpicture}[x=0.08cm,y=1.0cm]
% x-axis
\draw[thick] (0,-0.5) -- (100,-0.5);
\foreach \x in {0,20,40,60,80,100} {
    \draw (\x,-0.5) -- (\x,-0.65);
    \node[below] at (\x,-0.65) {\large \x};
}
\node at (50,-1.6) {\large Ratio (\%)};
% y-axis
\draw[thick] (0,-0.5) -- (0,5.5);
% y labels
\node[left] at (0,0) {\large Raon};
\node[left] at (0,1) {\large Qwen3};
\node[left] at (0,2) {\large GPT-Audio};
\node[left] at (0,3) {\large Gemma-3n};
\node[left] at (0,4) {\large Voxtral};
\node[left] at (0,5) {\large Gemini-Flash};
% ===== Raon =====
\fill[speechfollow] (0,-0.2) rectangle (36,0.4);
\fill[speechignore] (36,-0.2) rectangle (96,0.4);
\fill[speechmis]    (96,-0.2) rectangle (100,0.4);
\node at (18,0) {\normalsize 36};
\node at (66,0) {\normalsize 60};
\node at (98,0) {\normalsize 4};
% ===== Qwen3 =====
\fill[speechfollow] (0,0.7) rectangle (92.1,1.3);
\fill[speechignore] (92.1,0.7) rectangle (99.2,1.3);
\node at (46.05,1) {\normalsize 93};
\node at (95.65,1) {\normalsize 7};
% ===== GPT-4o =====
\fill[speechfollow] (0,1.7) rectangle (49.3,2.3);
\fill[speechignore] (49.3,1.7) rectangle (100,2.3);
\node at (24.65,2) {\normalsize 49};
\node at (74.65,2) {\normalsize 51};
% ===== Gemma-3n =====
\fill[speechfollow] (0,2.7) rectangle (57.1,3.3);
\fill[speechignore] (57.1,2.7) rectangle (100,3.3);
\node at (28.55,3) {\normalsize 57};
\node at (78.55,3) {\normalsize 43};
% ===== Voxtral =====
\fill[speechfollow] (0,3.7) rectangle (73.8,4.3);
\fill[speechignore] (73.8,3.7) rectangle (92.9,4.3);
\fill[speechmis]    (92.9,3.7) rectangle (100,4.3);
\node at (36.9,4) {\normalsize 74};
\node at (83.35,4) {\normalsize 19};
\node at (96.45,4) {\normalsize 7};
% ===== Gemini-Flash =====
\fill[speechfollow] (0,4.7) rectangle (63.7,5.3);
\fill[speechignore] (63.7,4.7) rectangle (100,5.3);
\node at (31.85,5) {\normalsize 64};
\node at (81.85,5) {\normalsize 36};
% ===== legend (horizontal layout) =====
\fill[speechfollow] (-20,-2.2) rectangle (-16,-1.9);
\node[right] at (-15,-2.05) {\large Followed speech};
\fill[speechignore] (28,-2.2) rectangle (32,-1.9);
\node[right] at (33,-2.05) {\large Ignored speech};
\fill[speechmis] (70,-2.2) rectangle (74,-1.9);
\node[right] at (75,-2.05) {\large Misaligned};
\end{tikzpicture}
}
\caption{Model behavior under correct speech contexts on the SCA-QA benchmark. The bar chart reports results for the K-history (before Chosun) subset.}
\label{fig:alignment_behavior_correct_speech}
\end{figure}

\paragraph{PA-QA.}
To analyze whether models attend to relevant speech evidence, we computed the evidence attention ratio (EAR) (Appendix~\ref{app:attention_analysis}) using teacher-forced generation, where the model-generated answer tokens are
appended to the prompt and processed in a single forward pass to extract attention weights. We measured the attention from the answer token to each speech sentence.
As shown in Figure~\ref{fig:ear_comparison}, EAR for the evidence sentence (located at the 47\% position of the story) exceeded 1.0 in most layers, with strong peaks in
layers 39-45, while irrelevant sentences remained near 1.0 across all layers.
Overall, the evidence sentence showed a substantially higher average EAR than irrelevant sentences (1.68 vs.\ 0.99), suggesting that the model selectively attends to the ground-truth evidence even when it appears in the middle of long speech input.
Notably, this analysis is enabled by the explicit evidence annotations in PA-QA, which allow fine-grained examination of whether a model's attention aligns with the intended evidence positions.

\begin{figure}[t]
\centering
\begin{tikzpicture}
\begin{axis}[
    width=\columnwidth,
    height=4cm,
    xlabel={Layer},
    ylabel={EAR},
    xlabel style={font=\small},
    ylabel style={font=\small},
    ymin=0, ymax=6.8,
    xmin=-1, xmax=48,
    xtick={0,8,16,24,32,40,47},
    ytick={0,2,4,6},
    minor y tick num=1,
    ymajorgrids=true,
    yminorgrids=true,
    major grid style={gray!20, thin},
    minor grid style={gray!10, very thin},
    tick label style={font=\footnotesize},
    legend style={
        at={(0.03,0.95)},
        anchor=north west,
        font=\footnotesize,
        draw=none,
        fill=white,
        fill opacity=0.85,
        text opacity=1,
        inner sep=3pt,
        row sep=2pt,
    },
    axis lines=left,
    axis line style={gray!50},
    every tick/.style={gray!50},
    axis on top,
]

% ---- Shaded area under irrelevant (red tint) ----
\addplot[red!8, opacity=0.5, forget plot, draw=none]
    coordinates {
        (0,1.011)  (1,1.005)  (2,1.037)  (3,1.004)
        (4,1.046)  (5,1.016)  (6,1.028)  (7,1.024)
        (8,0.957)  (9,1.012)  (10,0.993) (11,0.968)
        (12,0.932) (13,0.980) (14,0.988) (15,0.919)
        (16,0.985) (17,0.964) (18,0.967) (19,0.985)
        (20,0.939) (21,0.988) (22,1.011) (23,1.069)
        (24,1.035) (25,1.085) (26,0.982) (27,0.965)
        (28,0.958) (29,0.988) (30,0.952) (31,0.970)
        (32,1.023) (33,0.992) (34,1.102) (35,1.132)
        (36,1.102) (37,1.094) (38,0.974) (39,0.906)
        (40,0.794) (41,0.960) (42,0.983) (43,1.011)
        (44,0.993) (45,0.803) (46,0.952) (47,1.033)
        (47,0) (0,0)
    } \closedcycle;

% ---- Shaded gap: evidence(forward) -> irrelevant(backward) ----
\addplot[blue!12, opacity=0.7, forget plot, draw=none]
    coordinates {
        (0,0.860)  (1,0.801)  (2,0.992)  (3,0.733)
        (4,0.987)  (5,0.920)  (6,1.344)  (7,1.051)
        (8,0.972)  (9,1.133)  (10,1.371) (11,1.169)
        (12,1.781) (13,1.754) (14,1.619) (15,1.766)
        (16,1.889) (17,1.663) (18,1.446) (19,1.382)
        (20,1.498) (21,1.097) (22,0.947) (23,0.784)
        (24,0.910) (25,0.648) (26,1.753) (27,1.866)
        (28,2.087) (29,1.378) (30,1.877) (31,1.358)
        (32,0.985) (33,1.502) (34,1.124) (35,0.821)
        (36,1.617) (37,1.518) (38,2.010) (39,3.997)
        (40,5.906) (41,2.659) (42,3.761) (43,2.060)
        (44,1.486) (45,5.588) (46,2.660) (47,1.051)
        % -- reverse irrelevant --
        (47,1.033) (46,0.952) (45,0.803) (44,0.993)
        (43,1.011) (42,0.983) (41,0.960) (40,0.794)
        (39,0.906) (38,0.974) (37,1.094) (36,1.102)
        (35,1.132) (34,1.102) (33,0.992) (32,1.023)
        (31,0.970) (30,0.952) (29,0.988) (28,0.958)
        (27,0.965) (26,0.982) (25,1.085) (24,1.035)
        (23,1.069) (22,1.011) (21,0.988) (20,0.939)
        (19,0.985) (18,0.967) (17,0.964) (16,0.985)
        (15,0.919) (14,0.988) (13,0.980) (12,0.932)
        (11,0.968) (10,0.993) (9,1.012)  (8,0.957)
        (7,1.024)  (6,1.028)  (5,1.016)  (4,1.046)
        (3,1.004)  (2,1.037)  (1,1.005)  (0,1.011)
    } \closedcycle;

% ---- Irrelevant line ----
\addplot[
    color={rgb,255:red,231;green,76;blue,60},
    line width=1.0pt,
    smooth, tension=0.3,
] coordinates {
    (0,1.011)  (1,1.005)  (2,1.037)  (3,1.004)
    (4,1.046)  (5,1.016)  (6,1.028)  (7,1.024)
    (8,0.957)  (9,1.012)  (10,0.993) (11,0.968)
    (12,0.932) (13,0.980) (14,0.988) (15,0.919)
    (16,0.985) (17,0.964) (18,0.967) (19,0.985)
    (20,0.939) (21,0.988) (22,1.011) (23,1.069)
    (24,1.035) (25,1.085) (26,0.982) (27,0.965)
    (28,0.958) (29,0.988) (30,0.952) (31,0.970)
    (32,1.023) (33,0.992) (34,1.102) (35,1.132)
    (36,1.102) (37,1.094) (38,0.974) (39,0.906)
    (40,0.794) (41,0.960) (42,0.983) (43,1.011)
    (44,0.993) (45,0.803) (46,0.952) (47,1.033)
};
\addlegendentry{Irrelevant}

% ---- Evidence line ----
\addplot[
    color={rgb,255:red,52;green,120;blue,198},
    line width=1.2pt,
    smooth, tension=0.3,
] coordinates {
    (0,0.860)  (1,0.801)  (2,0.992)  (3,0.733)
    (4,0.987)  (5,0.920)  (6,1.344)  (7,1.051)
    (8,0.972)  (9,1.133)  (10,1.371) (11,1.169)
    (12,1.781) (13,1.754) (14,1.619) (15,1.766)
    (16,1.889) (17,1.663) (18,1.446) (19,1.382)
    (20,1.498) (21,1.097) (22,0.947) (23,0.784)
    (24,0.910) (25,0.648) (26,1.753) (27,1.866)
    (28,2.087) (29,1.378) (30,1.877) (31,1.358)
    (32,0.985) (33,1.502) (34,1.124) (35,0.821)
    (36,1.617) (37,1.518) (38,2.010) (39,3.997)
    (40,5.906) (41,2.659) (42,3.761) (43,2.060)
    (44,1.486) (45,5.588) (46,2.660) (47,1.051)
};
\addlegendentry{Evidence}

\end{axis}
\end{tikzpicture}
\caption{Per-layer EAR of the evidence sentence vs.\ irrelevant sentences (details of the sample are provided in Appendix~\ref{sec:ear_example}). The model allocates substantially more attention to the evidence in deeper layers (L39--L45).}
\label{fig:ear_comparison}
\end{figure}

\section{Conclusion}
In this paper, we introduce KoALa-Bench, a benchmark for evaluating Korean speech understanding in audio language models. KoALa-Bench includes ASR, ST, SQA, SIF, and two speech faithfulness tasks, SCA-QA and PA-QA, which evaluate modality and positional faithfulness to speech input with dedicated metrics. To assess Korea-specific knowledge, we incorporate KCSAT and crawled Korean-culture datasets. We evaluate both white-box and black-box models under clean and noisy conditions. 
% Our benchmark, code, and leaderboard are publicly available at \url{https://ksbench.github.io/Korean-Benchmark/}.

\section*{Limitations}
While KoALa-Bench provides a comprehensive evaluation framework for Korean speech understanding, it has two limitations. First, the benchmark is currently limited to Korean, leaving other non-English languages unexplored. Nevertheless, we believe our benchmark serves as an important step toward multilingual speech evaluation beyond English. Second, while the ASR task data and KCSAT listening questions consist of real human speech, the remaining datasets rely on TTS-synthesized speech, which may not fully capture the diversity of natural spontaneous speech. We leave the extension to real-world speech data and other languages as future work.

\section*{Ethical Considerations}
We construct KoALa-Bench from publicly available data while complying with applicable copyright and licensing requirements. All data are used for non-commercial academic research, and we do not redistribute raw copyrighted content, providing only processed annotations and metadata when permitted. We use the GPT and Gemini APIs for evaluation, and the GPT API is additionally used for dataset construction.
The dataset may inherit cultural and domain-specific biases from the source data, and evaluated models may exhibit hallucinations or unintended behaviors. This benchmark is intended for analyzing model behavior, particularly reliance on speech context, and is not designed for safety-critical use. We encourage responsible use and transparent reporting of model limitations.

\section*{Acknowledgments}
This work was partly supported by the National Research Foundation of Korea (NRF) grant funded by the Korea government (MSIT) (RS-2025-00553785) and the Institute of Information \& Communications Technology Planning \& Evaluation (IITP) grant funded by the Korean government (MSIT) (RS-2021-II211341, Artificial Intelligence Graduate School Program of Chung-Ang University). This work was conducted as part of an industry-academia research collaboration and was supported by Upstage. This research was also supported by the MSIT (Ministry of Science and ICT), Korea, under the Top-Tier AI Global HRD invitation program (RS-2025-25461932) supervised by the IITP (Institute for Information \& Communications Technology Planning \& Evaluation).

\bibliography{custom}
\appendix

\label{sec:appendix}
\section*{Appendices}
\section{Data Construction Details}
This section details the dataset construction and evaluation setup of KoALa-Bench. 
We describe the noise augmentation protocol, human annotation guidelines, representative examples from each evaluation subset, the SCA-QA construction pipeline, and the two-stage evidence grounding procedure for PA-QA.

\subsection{Details of Noise Injection for Robustness Experiments}
\label{sec:noise_dataset_injection}
% noise 관련 작성하기
\begin{table}[h]
\centering
\footnotesize
\setlength{\tabcolsep}{6pt}
\renewcommand{\arraystretch}{1.5}
\begin{tabular}{l|c}
\toprule\hline
\cellcolor{softgray}\textbf{Dataset} & \cellcolor{softgray}\textbf{\# Samples} \\
\hline\hline
Traffic & 6,292 \\
Environmental & 13,300 \\
Other & 8,000\\
\hline
\bottomrule
\end{tabular}
\caption{Types of Environmental Noise Used for Noise Augmentation.}
\label{tab:noise_dataset}
\end{table}

To evaluate robustness under acoustic degradation, we augmented the speech samples using both real environmental noise from the SELECTSTAR\footnote{\url{https://selectstar.ai/}} noise dataset and additional synthetic noise.
The SELECTSTAR dataset contains diverse environmental sounds such as traffic noise and ambient location noise and the detailed noise information is provided in Table~\ref{tab:noise_dataset}.
In addition to real-world noise, we also introduced synthetic noise types, including white noise, pink noise, brown noise, electrical hum, and impulse noise, to simulate a wider range of acoustic disturbances.
All noise signals were mixed with speech at randomly sampled signal-to-noise ratio (SNR) levels to generate the noise-augmented evaluation samples.

\subsection{Human annotation}
\label{app:human_annotation}
% We conduct a binary human evaluation to measure TTS naturalness. 
% Seven AI-major annotators rate each synthesized sample as either natural (1) or unnatural/awkward (0), following the guidelines in Table~\ref{tab:annotation_guidelines}.
We conduct a binary human evaluation to filter out unnatural TTS samples. The synthesized dataset is evenly divided among seven annotators, all native Korean speakers majoring in AI, where each annotator is responsible for reviewing their assigned partition. Each sample is rated as either natural (1) or unnatural/awkward (0) following the guidelines in Table~\ref{tab:annotation_guidelines}. Samples rated as unnatural are discarded from the final dataset. All annotators participated voluntarily as part of a collaborative research effort.
% \begin{table*}[t!]
% \centering
% \begin{tabular}{ll}
% \toprule
% \textbf{Item} & \textbf{Description} \\
% \midrule
% Task & TTS naturalness evaluation \\
% Annotators & 7 AI-major undergraduate/graduate students \\
% Recruitment & Voluntary participants from our research lab \\
% Compensation & No additional compensation provided \\
% Consent & All annotators were informed of data usage and provided consent \\
% \midrule
% \multicolumn{2}{l}{\textbf{Instruction}} \\
% \midrule
% \multicolumn{2}{p{0.9\textwidth}}{%
% TTS 합성이 자연스럽다면 1점, 그렇지 않고 부자연스럽거나 내용이 어색하다면 0점을 작성해주세요. \newline
% \textit{(Score 1 if the synthesized speech sounds natural; score 0 if it sounds unnatural or the content is awkward.)}} \\
% \midrule
% Score 1 & Natural-sounding synthesized speech \\
% Score 0 & Unnatural speech or awkward content \\
% \bottomrule
% \end{tabular}

% \caption{Human annotation guidelines for TTS quality evaluation.}
% \label{tab:annotation_guidelines}
% \end{table*}

\begin{table}[h]
\centering
\small
\setlength{\tabcolsep}{4pt}
\renewcommand{\arraystretch}{1.2}
\begin{tabularx}{\columnwidth}{@{}p{0.28\columnwidth}X@{}}
\toprule
\textbf{Item} & \textbf{Description} \\
\midrule
Task & TTS naturalness evaluation \\
Annotators & 7 AI-major undergraduate/graduate students \\
Recruitment & Voluntary participants from our research lab \\
Compensation & No additional compensation provided \\
Consent & All annotators were informed of data usage and provided consent \\
\midrule
\multicolumn{2}{@{}l}{\textbf{Instruction}} \\
\midrule
\multicolumn{2}{@{}p{\columnwidth}@{}}{%
TTS 합성이 자연스럽다면 1점, 그렇지 않고 부자연스럽거나 내용이 어색하다면 0점을 작성해주세요. \newline
\textit{(Score 1 if the synthesized speech sounds natural; score 0 if it sounds unnatural or the content is awkward.)}} \\
\midrule
Score 1 & Natural-sounding synthesized speech \\
Score 0 & Unnatural speech or awkward content \\
\bottomrule
\end{tabularx}
\caption{Human annotation guidelines for TTS quality evaluation.}
\label{tab:annotation_guidelines}
\end{table}

\subsection{Dataset Examples}
\label{sec:dataset_examples}
% This section provides representative examples from three evaluation datasets used in KoALa-Bench: KCSAT, SCA-QA, and PA-QA. Section~\ref{sec:kcsat_example} presents an example from the KCSAT dataset. Each example consists of a long-form speech context in the form of a dialogue, a multiple-choice question, and the corresponding answer. Section~\ref{sec:sca_qa_example} presents examples from the SCA-QA dataset across three domains: K-history, K-sports, and K-pop, respectively. Each example demonstrates how original speech contexts are modified through entity replacement to construct counterfactual pairs, where the keyword column indicates the entity used for replacement. Section~\ref{sec:pa_qa_example} shows an example from the PA-QA dataset. Each example includes evidence sentences mapped to their relative positions within the speech context, enabling fine-grained analysis of where the supporting information appears.
This section presents representative examples from the three evaluation datasets in KoALa-Bench: KCSAT, SCA-QA, and PA-QA. Section~\ref{sec:kcsat_example} shows a KCSAT example consisting of a long-form dialogue-style speech context, a multiple-choice question, and its answer. Section~\ref{sec:sca_qa_example} provides SCA-QA examples from K-history, K-sports, and K-pop, illustrating how entity replacement modifies original speech contexts to construct counterfactual pairs; the keyword column denotes the replacement entity. Section~\ref{sec:pa_qa_example} presents a PA-QA example with evidence sentences mapped to their relative positions in the speech context, supporting fine-grained analysis of where answer-relevant information appears.

\begin{table*}[t!]
\centering
\footnotesize
\setlength{\tabcolsep}{4pt}
\renewcommand{\arraystretch}{1.4}
\begin{tabular}{p{0.10\textwidth}|p{0.86\textwidth}}
\toprule\hline
\cellcolor{softgray}\textbf{Field} & \cellcolor{softgray}\textbf{Content} \\
\hline
\cellcolor{softgray}\textbf{Speech} \newline \textbf{Context} & \textbf{진행자} : 안녕하세요? 전문가와 함께 하는 삶의 이야기 시간입니다. 오늘은 건축가이신 김 선생님을 모시고 이야기를 들어 볼까 합니다. 안녕하세요? 선생님. \newline \textit{(\textbf{Host}: Hello? Welcome to Life Stories with Experts. Today, we have architect Mr. Kim with us to share his thoughts. Welcome, Mr. Kim.)} \newline\newline \textbf{김 선생님} : 네, 안녕하세요? 오늘은 제가 그동안 건축 분야에서 일하면서 느꼈던 계단과 우리네 삶의 의미에 대해 이야기를 해 보겠습니다. 누워서 세상을 보던 아이가 기기 시작하면서 보는 세상은, 이전과는 다른 새로운 세계입니다. 아이가 첫발을 내디디며 느끼는 것도 달라진 높이에서 보는 기쁨이죠. 일상적인 삶의 공간에 위치하고 있는 계단도 이런 즐거움과 삶의 의미를 전해 주죠. \newline \textit{(\textbf{Mr. Kim}: Yes, hello. Today, I would like to talk about stairs and the meaning they hold in our lives, based on my experience working in architecture. When a baby who once saw the world lying down begins to crawl, the world becomes an entirely new place. The joy a child feels when taking their first steps also comes from seeing things at a different height. Stairs, situated in our everyday living spaces, convey this same kind of joy and meaning in life.)} \newline\newline \textbf{진행자} : 네. 그렇다면 계단이 우리 삶에서 갖는 의미는 어떤 것인가요? \newline \textit{(\textbf{Host}: I see. Then, what meaning do stairs hold in our lives?)} \newline\newline \textbf{김 선생님} : 아래층과 위층은 들어오는 빛, 경치, 심지어는 냄새까지도 다른 공간입니다. 계단은 이렇게 분리된 두 공간을 이어 주고 있습니다. 계단을 오르내리면서 서로 다른 차원의 공간과 연결될 수 있는 거지요. 이렇게 우리는 계단을 통해 새로운 세계와 교류하게 되고, 달라진 높이에서 세상을 새롭게 조망할 수 있게 되는 것이죠. 결국 인간은 계단을 통해 \newline \textit{(\textbf{Mr. Kim}: The lower and upper floors are different spaces in terms of light, scenery, and even smell. Stairs connect these two separate spaces. By going up and down the stairs, we can be connected to spaces of different dimensions. In this way, through stairs, we come to interact with a new world and gain a fresh perspective on the world from a different height. Ultimately, through stairs, humans...)} \\
\hline
\cellcolor{softgray}\textbf{Question} & 방송의 마지막에 이어질 말로 가장 적절한 것은? \newline \textit{(What is the most appropriate statement to follow at the end of the broadcast?)} \\
\hline
\cellcolor{softgray}\textbf{Choices} & (1) 세상에 대한 긍정적인 시선을 갖게 됩니다. \textit{(We come to have a positive outlook on the world.)} \newline (2) 어려운 이웃을 배려하는 마음을 배우게 됩니다. \textit{(We learn to care for neighbors in need.)} \newline (3) 새로운 세상과 만나고 소통하는 기쁨을 얻게 됩니다. \textit{(We gain the joy of encountering and communicating with a new world.)} \newline (4) 끊임없는 도전만이 성공을 위한 열쇠라는 것을 알게 됩니다. \textit{(We learn that only constant challenge is the key to success.)} \newline (5) 일의 성패는 스스로의 노력에 달려 있다는 것을 배우게 됩니다. \textit{(We learn that the success or failure of a task depends on one's own effort.)} \\
\hline
\cellcolor{softgray}\textbf{Answer} & (3) \\
\hline
\bottomrule
\end{tabular}
\caption{
A representative example from the KCSAT subset. 
The example requires selecting the most appropriate continuation by reasoning over the dialogue context.
}
\label{tab:kcsat_datasets}
\end{table*}

\subsubsection{KCSAT}
\label{sec:kcsat_example}
% Table~\ref{tab:kcsat_datasets} shows a subset of examples from the KCSAT dataset, which is derived from the listening section of the Korean College Scholastic Ability Test. Each example consists of a long-form speech context in the form of a dialogue, a multiple-choice question, and the corresponding answer. The speech contexts are segmented from the original audio files into individual question-level instances.
Table~\ref{tab:kcsat_datasets} presents a representative example from the KCSAT subset, which is derived from the listening section of the Korean College Scholastic Ability Test (KCSAT).\footnote{Official materials are available at \url{https://www.ebsi.co.kr/}} We use all listening examinations administered between 2006 and 2012, 
after which the Korean listening section was discontinued.Each instance is constructed at the question level by segmenting the listening audio into a long-form speech context paired with a multiple-choice question and its ground-truth answer. The questions cover diverse discourse types, including lectures, presentations, debates, interviews, and multi-turn dialogues, and require various reasoning skills such as identifying the speaker's intent, inferring the most appropriate continuation of an incomplete utterance, and comprehending the main argument or factual details from the speech context. In the example shown in Table~\ref{tab:kcsat_datasets}, the final utterance is incomplete, and the model must select the most appropriate continuation from five candidates. The correct answer is (3), since the preceding dialogue states that stairs connect separate spaces, allow people to interact with a new world, and provide a new perspective from a different height. Thus, the example requires discourse-level inference over the dialogue context rather than matching a single explicit phrase in the speech.

\begin{table*}[t!]
\centering
\footnotesize
\setlength{\tabcolsep}{4pt}
\renewcommand{\arraystretch}{1.4}
\begin{tabular}{p{0.1\textwidth}|p{0.42\textwidth}|p{0.42\textwidth}}
\toprule\hline
\cellcolor{softgray} & \cellcolor{softgray}\textbf{Original} & \cellcolor{softgray}\textbf{Counterfactual} \\
\hline
\cellcolor{softgray}\textbf{Speech} \newline \textbf{Context} & 세종대왕 동상은 대한민국 서울특별시 종로구 세종로 광화문 광장에 있는 세종대왕 동상이다. 세종대왕 동상은 홍익대학교 교수이자 조각가인 김영원이 설계한 것으로, 2009년 건립되었다. \newline \textit{(Sejong the Great Statue is a statue of Sejong the Great located in Gwanghwamun Square, Sejong-ro, Jongno-gu, Seoul, Republic of Korea. The statue was designed by sculptor Kim Young-won, a professor at Hongik University, and was erected in 2009.)} & 이순신 동상은 대한민국 서울특별시 종로구 세종로 광화문 광장에 있는 동상이다. 이순신 동상은 홍익대학교 교수이자 조각가인 김영원이 설계한 것으로, 2009년 건립되었다. \newline \textit{(Yi Sun-sin Statue is a statue of Yi Sun-sin located in Gwanghwamun Square, Sejong-ro, Jongno-gu, Seoul, Republic of Korea. The statue was designed by sculptor Kim Young-won, a professor at Hongik University, and was erected in 2009.)} \\
\hline
\cellcolor{softgray}\textbf{Question} & \multicolumn{2}{p{0.82\textwidth}}{대한민국 서울특별시 종로구 세종로 광화문 광장에 위치하며, 홍익대학교 교수인 김영원이 설계하고 2009년에 건립된 동상의 이름은 무엇인가? \newline \textit{(What is the name of the statue located at Gwanghwamun Square in Sejong-ro, Jongno-gu, Seoul, Republic of Korea, which was designed by Kim Young-won, a professor at Hongik University, and erected in 2009?)}} \\
\hline
\cellcolor{softgray}\textbf{Keyword} & 세종대왕 \newline \textit{(Sejong the Great)} & 이순신 \newline \textit{(Yi Sun-sin)} \\
\hline
\cellcolor{softgray}\textbf{Answer} & 세종대왕 동상 \newline \textit{(Statue of Sejong the Great)} & 이순신 동상 \newline \textit{(Statue of Yi Sun-sin)} \\
\hline
\bottomrule
\end{tabular}
\caption{
A representative K-history example from SCA-QA. 
The counterfactual context replaces the target entity from Sejong the Great to Yi Sun-sin, changing the speech grounded answer from the Sejong the Great Statue to the Yi Sun-sin Statue.
}
\label{tab:sca_qa_datasets_khistory}
\end{table*}

\begin{table*}[t!]
\centering
\footnotesize
\setlength{\tabcolsep}{4pt}
\renewcommand{\arraystretch}{1.4}
\begin{tabular}{p{0.1\textwidth}|p{0.42\textwidth}|p{0.42\textwidth}}
\toprule\hline
\cellcolor{softgray} & \cellcolor{softgray}\textbf{Original} & \cellcolor{softgray}\textbf{Counterfactual} \\
\hline
\cellcolor{softgray}\textbf{Speech} \newline \textbf{Context} & 유창혁은 대한민국의 바둑 기사이다. 1966년에 서울특별시에서 태어났고, 1979년에 어린이 국수전 3연패 학초배 우승하였다. 1984년 입단 이후 신인 대회 수상을 시작으로 세계대회 우승 6회 등 세계대회 그랜드슬램을 달성한 바둑 기사이다. \newline \textit{(Yoo Chang-hyuk is a Korean professional Go player. He was born in Seoul in 1966 and won three consecutive Children's Guksu titles and the Hakcho Cup in 1979. After turning professional in 1984, he began winning rookie tournaments and went on to achieve a Grand Slam in international competitions, with six world championship titles.)} & 이세돌은 대한민국의 바둑 기사이다. 1966년에 서울특별시에서 태어났고, 1979년에 어린이 국수전 3연패 학초배 우승하였다. 1984년 입단 이후 신인 대회 수상을 시작으로 세계대회 우승 6회 등 세계대회 그랜드슬램을 달성한 바둑 기사이다. \newline \textit{(Lee Sedol is a Korean professional Go player. He was born in Seoul in 1966 and won three consecutive Children's Guksu titles and the Hakcho Cup in 1979. After turning professional in 1984, he began winning rookie tournaments and went on to achieve a Grand Slam in international competitions, with six world championship titles.)} \\
\hline
\cellcolor{softgray}\textbf{Question} & \multicolumn{2}{p{0.82\textwidth}}{대한민국의 바둑 기사로, 1966년에 서울특별시에서 태어나 어린이 국수전 3연패와 학초배 우승을 차지한 후, 1984년 입단하여 세계대회에서 6회 우승을 기록한 인물은 누구인가? \newline \textit{(Who is the Korean professional Go player, born in Seoul in 1966, who won three consecutive Children's Guksu titles and the Hakcho Cup, turned professional in 1984, and went on to win six world championship titles?)}} \\
\hline
\cellcolor{softgray}\textbf{Keyword} & 유창혁 \newline \textit{(Yoo Chang-hyuk)} & 이세돌 \newline \textit{(Lee Sedol)} \\
\hline
\cellcolor{softgray}\textbf{Answer} & 유창혁 \newline \textit{(Yoo Chang-hyuk)} & 이세돌 \newline \textit{(Lee Sedol)} \\
\hline
\bottomrule
\end{tabular}
\caption{
A representative K-sports example from SCA-QA. 
The counterfactual context replaces the target entity from Yoo Chang-hyuk to Lee Sedol, requiring the answer to change according to the entity stated in the speech context.
}
\label{tab:sca_qa_datasets_ksports}
\end{table*}

\begin{table*}[t!]
\centering
\footnotesize
\setlength{\tabcolsep}{4pt}
\renewcommand{\arraystretch}{1.4}
\begin{tabular}{p{0.1\textwidth}|p{0.42\textwidth}|p{0.42\textwidth}}
\toprule\hline
\cellcolor{softgray} & \cellcolor{softgray}\textbf{Original} & \cellcolor{softgray}\textbf{Counterfactual} \\
\hline
\cellcolor{softgray}\textbf{Speech} \newline \textbf{Context} & 다비치는 2008년 2월 22일 데뷔한 대한민국의 여성 보컬 듀오이다. 2008년, 정규 1집 Amaranth를 발매하며 성공적으로 데뷔했다. \newline \textit{(Davichi is a South Korean female vocal duo that debuted on February 22, 2008. They made a successful debut in 2008 with the release of their first studio album, Amaranth.)} & 마마무는 2008년 2월 22일 데뷔한 대한민국의 여성 보컬 듀오이다. 2008년, 정규 1집 Amaranth를 발매하며 성공적으로 데뷔했다. \newline \textit{(MAMAMOO is a South Korean female vocal duo that debuted on February 22, 2008. They made a successful debut in 2008 with the release of their first studio album, Amaranth.)} \\
\hline
\cellcolor{softgray}\textbf{Question} & \multicolumn{2}{p{0.82\textwidth}}{2008년 2월 22일에 데뷔하여 정규 1집 Amaranth를 발매하며 성공적으로 출발한 대한민국의 여성 보컬 듀오는 무엇인가? \newline \textit{(What is the South Korean female vocal duo that debuted on February 22, 2008, and made a successful start with the release of their first studio album, Amaranth?)}} \\
\hline
\cellcolor{softgray}\textbf{Keyword} & 다비치 \newline \textit{(Davichi)} & 마마무 \newline \textit{(MAMAMOO)} \\
\hline
\cellcolor{softgray}\textbf{Answer} & 다비치 \newline \textit{(Davichi)} & 마마무 \newline \textit{(MAMAMOO)} \\
\hline
\bottomrule
\end{tabular}
\caption{
A representative K-pop example from SCA-QA. 
The counterfactual context replaces the target entity from Davichi to MAMAMOO, changing the correct answer while preserving the overall question structure.
}
\label{tab:sca_qa_datasets_kpop}
\end{table*}

\begin{table*}[t!]
\centering
\footnotesize
\setlength{\tabcolsep}{4pt}
\renewcommand{\arraystretch}{1.4}
\begin{tabular}{p{0.10\textwidth}|p{0.85\textwidth}}
\toprule\hline
\cellcolor{softgray}\textbf{Field} & \cellcolor{softgray}\textbf{Content} \\
\hline
\cellcolor{softgray}\textbf{Speech} \newline \textbf{Context} & 동물원 밖은 평소보다 따뜻했다. 광대는 분장을 한 채 땀을 흘리고 있었다. 하지만 그는 여전히 미소를 지으며 사람들을 웃게 만들었다. 그는 열심히 일했다. 누군가의 아들이 부탁해서 그는 종이 비행기를 던졌다. 작은 소녀가 그에게 ``이렇게 하면 더 예뻐 보일 거예요''라고 말하자, 그는 심지어 얼굴에 젤리를 발랐다. 정오에는 접시를 던지며 저글링을 했고, 오후 1시에는 앉아서 점심을 먹었다. 요리사는 광대가 좋아하는 스타일로 샐러드를 만들어 주었고, 광대는 기분 좋게 샐러드를 먹었다. 동물원을 방문한 사람들이 그를 가리키며 미소를 짓는 것을 보면서 그는 유명해진 기분을 느꼈다. 그날은 더웠지만, 사람들의 반응을 보면서 광대는 결국 좋은 하루였다고 생각했다. 얼마 지나지 않아 그는 다시 저글링을 하고 농담을 하며 일에 몰두했다. 심지어 누군가가 그의 가짜 꽃 냄새를 맡자 물을 뿌리기도 했다. \newline \textit{(It was unusually warm outside the zoo, and the clown was sweating in his costume. 
Still, he kept smiling and worked hard to entertain visitors. 
At a child's request, he threw a paper airplane, and when a little girl suggested it, he even put jelly on his face. 
He later juggled plates, ate his favorite salad for lunch, and felt pleased as visitors pointed at him and smiled. 
Although the day was hot, he thought it had been a good day and soon returned to juggling, joking, and spraying water from his fake flower.)}\\
\hline
\cellcolor{softgray}\textbf{Question} & 광대는 어떻게 그렇게 열심히 일할까요? \newline \textit{(How does the clown work so hard?)} \\
\hline
\cellcolor{softgray}\textbf{Choices} & (A) 그는 동물원에 머물렀다. \textit{(He stayed at the zoo.)} \newline (B) 그는 비행기 장난감을 던지고 얼굴에 젤리를 발랐다. \textit{(He threw a toy airplane and put jelly on his face.)} \newline (C) 그는 웃었다. \textit{(He laughed.)} \newline (D) 그는 밖에서 서 있었다. \textit{(He stood outside.)} \\
\hline
\cellcolor{softgray}\textbf{Evidence} & 누군가의 아들이 부탁해서 그는 종이 비행기를 던졌다. 작은 소녀가 그에게 ``이렇게 하면 더 예뻐 보일 거예요''라고 말하자, 그는 심지어 얼굴에 젤리를 발랐다. \newline \textit{(Someone's son asked, so he threw a paper airplane. When a little girl told him ``You'll look prettier if you do this,'' he even put jelly on his face.)} \\
\hline
\cellcolor{softgray}\textbf{Position} & Front, Front--middle \\
\hline
\cellcolor{softgray}\textbf{Answer} & (B) \\
\hline
\bottomrule
\end{tabular}
\caption{
A representative example from the PA-QA dataset. 
Each instance contains a long-form speech context, a multiple-choice question, the ground-truth answer, and evidence sentences annotated with their relative positions in the speech context.
}
\label{tab:pa_qa_datasets}
\end{table*}

\subsubsection{SCA-QA}
\label{sec:sca_qa_example}
% Table~\ref{tab:sca_qa_datasets_khistory}, \ref{tab:sca_qa_datasets_ksports}, and \ref{tab:sca_qa_datasets_kpop} show subsets of examples from the SCA-QA dataset across three domains: K-history, K-sports, and K-pop, respectively. Each example demonstrates how original speech contexts are modified through entity replacement to construct counterfactual pairs, where the keyword column indicates the entity used for replacement.

We collect 200 samples for each of four domains: pre-Joseon Korean history, post-Joseon Korean history, K-sports, and K-pop, yielding 800 initial samples. 
After human validation for counterfactual consistency, 374 samples are retained in the final SCA-QA dataset. 
Tables~\ref{tab:sca_qa_datasets_khistory}, \ref{tab:sca_qa_datasets_ksports}, and \ref{tab:sca_qa_datasets_kpop} show representative counterfactual examples from K-history, K-sports, and K-pop, respectively. 
Each example pairs an original speech context with a counterfactual version created by replacing the target entity while preserving the question structure. 
The keyword field specifies the entity that determines the speech-grounded answer.

\paragraph{K-history.}
Table~\ref{tab:sca_qa_datasets_khistory} shows a K-history example where the target entity is replaced from Sejong the Great to Yi Sun-sin, changing the answer from the Sejong the Great Statue to the Yi Sun-sin Statue according to the speech context.

\paragraph{K-sports.}
Table~\ref{tab:sca_qa_datasets_ksports} presents a K-sports example involving Korean professional Go players. The original context identifies Yoo Chang-hyuk, while the counterfactual context replaces him with Lee Sedol while preserving the surrounding biographical description. Thus, the model must update its answer based on the speech context rather than relying on prior knowledge about the described attributes.

\paragraph{K-pop.} Table~\ref{tab:sca_qa_datasets_kpop} shows a K-pop example where the described music group is changed from Davichi to MAMAMOO. Although the debut date and album information remain unchanged, the correct answer must follow the group name explicitly stated in the speech. This tests whether the model grounds its response in the provided speech context when it conflicts with potentially memorized associations.

Overall, these examples show that SCA-QA evaluates whether models ground their answers in the provided speech context rather than relying on parametric knowledge.

\begin{algorithm*}[!t]
\caption{SCA-QA Dataset Construction Pipeline}
\label{alg:sca_qa_pipeline}
\KwIn{Wikipedia passages $\mathcal{D}$, LLM client $\mathcal{G}$, TTS engine $\mathcal{T}$}
\KwOut{Multiple-choice QA samples with counterfactual speech contexts}

\textbf{Stage 1: Question and Answer Generation}\;
\For{passage $x \in \mathcal{D}$}{
  Generate question $q$ and answer $a$ from $x$ using $\mathcal{G}$ with 3-shot prompting\;
}

\textbf{Stage 2: Entity Replacement}\;
\For{each $(x, q, a)$}{
  Generate a new entity $a'$ of the same type as $a$ using $\mathcal{G}$, ensuring $a' \notin x$\;
}

\textbf{Stage 3: Sentence Transformation}\;
\For{each $(x, a, a')$}{
  Replace all occurrences of $a$ with $a'$ in $x$ to produce counterfactual context $x'$ using $\mathcal{G}$\;
}

\textbf{Stage 4: TTS Text Normalization}\;
\For{each $x'$}{
  Normalize numerals and non-Korean characters into spoken Korean forms using $\mathcal{G}$\;
  Synthesize speech from the normalized text using $\mathcal{T}$\;
}

\textbf{Stage 5: Distractor Generation}\;
\For{each $(a, a')$}{
  Generate two distractors $\{d_1, d_2\}$ of the same entity type using $\mathcal{G}$\;
  Construct choices $\mathcal{C} = \{a', a, d_1, d_2\}$ and shuffle; record correct index\;
}

\textbf{Stage 6: Answer Verification}\;
\For{each $(x', q, \mathcal{C}, a')$}{
  Query $\mathcal{G}$ with $x'$, $q$, and $\mathcal{C}$\;
  Discard sample if $\mathcal{G}$'s response $\neq a'$\;
}
\end{algorithm*}

\subsubsection{PA-QA}
\label{sec:pa_qa_example}
Table~\ref{tab:pa_qa_datasets} shows an example from the PA-QA dataset, which evaluates position-aware question answering over long-form speech inputs. Each example consists of a speech context, a multiple-choice question, and the corresponding evidence sentences that support the correct answer. The evidence sentences are identified through the evidence grounding pipeline and mapped to their relative positions within the speech context, which is normalized and divided into four temporal segments: front (0--0.25), front-middle (0.25--0.5), middle-last (0.5--0.75), and last (0.75--1.0). In the example shown, the evidence spans the front and front-middle segments, indicating that the supporting information appears in the first half of the speech context. This example illustrates how PA-QA enables fine-grained analysis of whether LALMs can faithfully utilize information appearing at different locations within long-form speech inputs.

\subsection{SCA-QA Pipeline}

Algorithm~\ref{alg:sca_qa_pipeline} summarizes the construction procedure for SCA-QA. Given a Wikipedia passage $x$, we use the LLM client $\mathcal{G}$ with 3-shot prompting to generate a question--answer pair $(q,a)$. We then generate a counterfactual entity $a'$ that has the same semantic type as $a$ but does not appear in $x$, and replace all occurrences of $a$ with $a'$ to obtain a counterfactual context $x'$. The question $q$ is kept unchanged, so the correct answer becomes $a'$ only when the model grounds its prediction in the modified context rather than relying on the original factual association with $a$. The counterfactual context is normalized into spoken Korean form and synthesized into speech using the TTS engine $\mathcal{T}$. We construct a four-way multiple-choice set $\mathcal{C}$ from the counterfactual answer $a'$, the original answer $a$, and two same-type distractors generated by $\mathcal{G}$. To ensure sample validity, we retain only examples where $\mathcal{G}$ selects $a'$ when given $(x', q, \mathcal{C})$, filtering out ambiguous or inconsistent cases.

\begin{algorithm*}[!t]
\caption{Two-Stage Evidence Grounding: Text $\rightarrow$ Audio}
\label{alg:two_stage_grounding}
\KwIn{Samples $\mathcal{D}$, GPT client $\mathcal{G}$, ASR model $\mathcal{W}$}
\KwOut{Samples with evidence sentences, validity labels, text positions, and audio time spans}

\textbf{Stage 1: Text evidence grounding}\;
\For{$r \in \mathcal{D}$}{
  Extract context $x$, question $q$, and gold answer $a^\star$\;
  \If{$x=\emptyset$}{
    Set $\texttt{is\_valid}\leftarrow\texttt{False}$; continue\;
  }

  Split $x$ into sentences $\{s_i\}_{i=1}^{N}$ and record their text spans\;
  Construct a candidate evidence pool $\mathcal{P}$ using answer-based rule matching\;

  \eIf{$\mathcal{P}\neq\emptyset$}{
    Select initial evidence sentences $\mathcal{E}$ from $\mathcal{P}$, using $\mathcal{G}$ when candidate selection is ambiguous\;
  }{
    Use $\mathcal{G}$ to select evidence sentences from the sentence pool $\{s_i\}_{i=1}^{N}$\;
  }

  Verify $\mathcal{E}$ with $\mathcal{G}$ given $(q, a^\star, \mathcal{E})$\;
  \If{the verification indicates that additional evidence is needed}{
    Select an additional evidence sentence using $\mathcal{G}$ and update $\mathcal{E}$\;
  }

  Re-verify the final evidence set $\mathcal{E}$ with $\mathcal{G}$\;
  Set $\texttt{is\_valid}\leftarrow(\texttt{supported}\land\neg\texttt{need\_more})$\;

  Map each evidence sentence in $\mathcal{E}$ to its relative text position by normalizing its start position within $x$\;
  Save evidence sentences, text spans, relative positions, temporal segment labels, verification result, and $\texttt{is\_valid}$\;
}

\textbf{Stage 2: Audio evidence grounding}\;
$\mathcal{D}_v \leftarrow \{r \in \mathcal{D}\mid r.\texttt{is\_valid}=\texttt{True}\}$\;
\For{$r \in \mathcal{D}_v$}{
  Transcribe the audio with word-level timestamps using $\mathcal{W}$\;
  Build a text-to-time mapping from the transcript\;
  \For{each evidence sentence $s \in \mathcal{E}$}{
    Align $s$ to the transcript using exact matching followed by fuzzy matching\;
    Convert the aligned text span into an audio time span $[t_s, t_e]$; if unmatched, set \texttt{None}\;
  }
  Save the audio time spans for the evidence sentences\;
}
\end{algorithm*}

\subsection{Two Stage Evidence Grounding for PA-QA}

\label{sec:appendix_grounding}

Algorithm~\ref{alg:two_stage_grounding} summarizes our two-stage procedure for grounding answer-relevant evidence from text to audio. In the first stage, each sample is decomposed into its context $x$, question $q$, and gold answer $a^\star$. We split $x$ into sentences and identify candidate evidence sentences using answer-based rule matching, with $\mathcal{G}$ used to resolve ambiguous cases or to select evidence directly when no rule-based candidate is found. The selected evidence set $\mathcal{E}$ is then verified by $\mathcal{G}$ with respect to $(q,a^\star)$, and additional evidence is added when the answer is not fully supported. We retain only samples whose final evidence set is verified as sufficient, and record each evidence sentence with its text span, normalized relative position, and temporal segment label.
In the second stage, we map the verified text evidence to the audio signal. For each valid sample, the audio is transcribed using the ASR model $\mathcal{W}$ with word-level timestamps. Each evidence sentence in $\mathcal{E}$ is aligned to the transcript using exact matching followed by fuzzy matching, and the matched text span is converted into an audio time span $[t_s,t_e]$. This procedure produces samples with both text-level evidence positions and audio-level evidence spans, enabling position-aware analysis of whether models use the relevant part of the speech input.

\subsection{Quality Validation of Synthesized Speech and Counterfactual Annotations}
\label{app:quality_validation}

Since a substantial portion of KoALa-Bench relies on synthesized speech and on automatically constructed counterfactual contexts, we conduct additional validation of both components. Section~\ref{app:tts_quality} reports the intelligibility and perceptual quality of the synthesized speech, and Section~\ref{app:sca_qa_agreement} reports the inter-annotator agreement of the SCA-QA counterfactual contexts.

\subsubsection{TTS Quality Validation}
\label{app:tts_quality}

We validate the quality of the synthesized speech along two complementary axes: objective intelligibility and perceptual naturalness. For intelligibility, we transcribe the full synthesized portion of each task with Qwen3-ASR-1.7B, an ASR system independent of the TTS engine used for synthesis, and compute the character error rate (CER) against the source text. For perceptual quality, we conduct a mean opinion score (MOS) test in which ten native Korean listeners independently rate ten randomly sampled utterances per task (40 utterances in total) on a five-point scale. The samples are presented in randomized order and without task identity, so that ratings are not biased by knowledge of the underlying task. Speaker identity is randomly sampled from the Qwen3-TTS speaker pool, and every synthesized sample additionally undergoes the manual listening verification described in Appendix~\ref{app:human_annotation} prior to inclusion in the benchmark.

Table~\ref{tab:tts_quality} summarizes the results. SQA, SIF, and PA-QA all exhibit low CER, indicating that an independent ASR system transcribes the synthesized speech accurately. SCA-QA shows a higher CER of 12.04\%, which we attribute to the Korean culture-specific proper nouns that dominate this task (e.g., K-pop artist names and historical figures), as such entities are frequently mis-transcribed even for human recordings. The MOS results support this interpretation: all four tasks score above 4.0, and SCA-QA obtains 4.00 $\pm$ 0.18 despite its elevated CER. The gap between the two metrics therefore reflects surface-level transcription errors on proper nouns rather than degraded speech quality. We note that synthesized speech does not reproduce the disfluencies and prosodic variability of spontaneous speech; KoALa-Bench mitigates this by including real human recordings for the ASR and KCSAT subsets.

\begin{table}[h]
\centering
\small
\setlength{\tabcolsep}{6pt}
\renewcommand{\arraystretch}{1.15}
\begin{tabular}{@{}lcc@{}}
\toprule
\textbf{Task} & \textbf{CER (\%)} $\downarrow$ & \textbf{MOS} $\uparrow$ \\
\midrule
SQA    & 2.92  & 4.12 $\pm$ 0.19 \\
SIF    & 1.03  & 4.41 $\pm$ 0.16 \\
SCA-QA & 12.04 & 4.00 $\pm$ 0.18 \\
PA-QA  & 3.62  & 4.33 $\pm$ 0.17 \\
\bottomrule
\end{tabular}
\caption{Quality validation of the synthesized speech. CER (\%) is measured by Qwen3-ASR-1.7B over the full synthesized set of each task, and MOS (1--5) is rated by ten listeners on ten randomly sampled utterances per task. $\pm$ denotes the 95\% confidence interval.}
\label{tab:tts_quality}
\end{table}

\subsubsection{Annotation Consistency of SCA-QA}
\label{app:sca_qa_agreement}

The counterfactual contexts of SCA-QA are generated by substituting entities with an LLM (Section~\ref{sec:sca_qa_example}), which raises the question of whether the resulting contexts are grammatically well-formed and plausible as spoken input. As described in Section~\ref{sec:sca_qa_example}, every generated context already undergoes human verification, in which annotators discard grammatically unnatural or implausible replacements. To quantify the reliability of this verification, we additionally perform a cross-annotation study.

We randomly sample 50 examples from the K-pop domain, which contains the highest density of proper nouns and is therefore the most demanding case for entity substitution. Three annotators, all native Korean speakers, independently judge each example along three criteria: grammatical correctness of the substituted context, naturalness when read aloud, and plausibility of the counterfactual entity as a same-type replacement. The annotators work independently and are not shown each other's decisions. The three annotators achieve a Fleiss' $\kappa$ of 0.66, indicating substantial agreement \citep{landis1977measurement}. Table~\ref{tab:sca_qa_agreement} summarizes the setting. This result indicates that the acceptability of the generated counterfactual contexts is judged consistently across annotators, supporting the reliability of the verification stage in the SCA-QA construction pipeline.

\begin{table}[h]
\centering
\small
\setlength{\tabcolsep}{4pt}
\renewcommand{\arraystretch}{1.2}
\begin{tabularx}{\columnwidth}{@{}p{0.30\columnwidth}X@{}}
\toprule
\textbf{Item} & \textbf{Description} \\
\midrule
Target        & SCA-QA counterfactual contexts (K-pop domain) \\
Samples       & 50 randomly sampled examples \\
Annotators    & 3 native Korean speakers, annotating independently \\
Criteria      & Grammatical correctness, naturalness, plausibility of the replaced entity \\
Agreement     & Fleiss' $\kappa$ = 0.66 (substantial) \\
\bottomrule
\end{tabularx}
\caption{Cross-annotation setting and inter-annotator agreement for the SCA-QA counterfactual contexts.}
\label{tab:sca_qa_agreement}
\end{table}

% ============================================================
\section{Evaluation Details}
\label{sec:evaluation_appendix}
This section details the evaluation protocols used in KoALa-Bench. 
We describe the normalization pipeline for Korean ASR, formally define the SCF-score for measuring speech-context faithfulness in SCA-QA, and introduce the Evidence Attention Ratio (EAR) for analyzing whether models attend to answer-relevant evidence in long speech inputs.

% ============================================================
% TABLE — Concrete examples
% ============================================================

\begin{table*}[t]
\centering
\small
\setlength{\tabcolsep}{5pt}
\renewcommand{\arraystretch}{1.15}
\begin{tabular}{
    @{}
    l
    l
    r
    c
    >{\raggedright\arraybackslash}p{5.6cm}
    @{}
}
\toprule
\textbf{Ground Truth (GT)}
    & \textbf{Prediction}
    & \textbf{CER}
    &
    & \textbf{Explanation} \\
\midrule

%%% --- Group 1: Punctuation & space normalization ---
\addlinespace[2pt]
\rowcolor{softgray}
\multicolumn{5}{@{}l@{}}{%
    \hspace{4pt}\textbf{Punctuation \& space normalization}
    \textemdash\ Symbols removed, spaces collapsed%
} \\
\addlinespace[2pt]
안녕하세요!      & 안녕하세요     & 0\% & \cmark & Punctuation ``!'' removed from GT \\
정말요?          & 정말요         & 0\% & \cmark & Question mark ``?'' removed from GT \\
네, 알겠습니다.  & 네알겠습니다   & 0\% & \cmark & Comma and period removed \\
아... 그래요     & 아그래요       & 0\% & \cmark & Ellipsis and spaces removed \\
감사 합니다      & 감사합니다     & 0\% & \cmark & Spaces removed from both sides \\

%%% --- Group 4: Hangul GT ---
\addlinespace[5pt]
\rowcolor{softgray}
\multicolumn{5}{@{}l@{}}{%
    \hspace{4pt}\textbf{Hangul ground truth}
    \textemdash\ No number conversion applied%
} \\
\addlinespace[2pt]
오백 원 & 오백원 & 0\%    & \cmark & Spaces removed; Hangul preserved as-is \\
오백 원 & 오천원 & $>$ 0\% & \xmark & No variant generated; ``오백'' $\neq$ ``오천'' \\
세 잔   & 세잔   & 0\%    & \cmark & Spaces removed; Hangul preserved as-is \\
세 잔   & 삼잔   & $>$ 0\% & \xmark & No variant generated for Hangul GT \\

\bottomrule
\end{tabular}
\caption{%
    Examples of CER evaluation under the normalization pipeline.
    Group 1 shows punctuation and space handling;
    group 2 shows that Hangul ground truth is preserved as-is.
    \cmark\ = accepted (CER\,=\,0\%),
    \xmark\ = rejected (CER\,$>$ \,0\%).
}
\label{tab:cer-examples}
\end{table*}

\subsection{CER Normalization and Computation}

\paragraph{Normalization Protocol.}
We evaluate Korean ASR using character error rate (CER)
with a normalization pipeline tailored to the linguistic properties of Korean numerals.
In Korean ASR evaluation, surface-level discrepancies
between the ground-truth transcription and the model hypothesis
frequently arise due to punctuation and spacing differences.
ASR models often omit or insert punctuation that is
present (or absent) in the ground truth,
and Korean word spacing conventions are inherently ambiguous, as even native speakers may segment words inconsistently.
Na\"ively computing CER on such pairs would incur
false penalties despite semantically identical transcriptions.

\paragraph{Normalization Pipeline.}
To address this, before computing CER, we apply a normalization step to both
the reference and the hypothesis, where all punctuation,
special characters, and whitespace are removed,
retaining only Hangul characters and Arabic digits.
This prevents surface-level discrepancies, such as
punctuation insertion or omission and spacing inconsistencies,
from inflating CER, ensuring that semantically identical
transcriptions are not unfairly penalized.
Table~\ref{tab:cer-examples} shows concrete examples
covering the normalization rules.
\paragraph{CER Computation.}
CER is computed at the corpus level via micro-averaging:
\begin{equation}
    \text{CER}
    = \frac{\displaystyle\sum_{i=1}^{N} d(r_i,\, h_i)}
           {\displaystyle\sum_{i=1}^{N} |r_i|}
    \label{eq:cer}
\end{equation}
where $d(r_i, h_i)$ is the Levenshtein edit distance between
the normalized reference~$r_i$ and hypothesis~$h_i$,
and $|r_i|$ is the character length of the normalized reference.

\subsection{Evaluation Metric of SCA-QA}
In this section, we formally define SCF-score, which measures whether model predictions rely on the provided speech context rather than parametric knowledge.

Let $\mathcal{D} = \{(q_i, a_i^{\text{text}}, a_i^{\text{speech}}, s_i)\}_{i=1}^{N}$ denote the evaluation set, where $q_i$ is a textual question, $a_i^{\text{text}}$ is the ground-truth answer to $q_i$, $a_i^{\text{speech}}$ is the answer derived from the speech context $s_i$ that conflicts with $a_i^{\text{text}}$, and $f$ denotes the model under evaluation. We define
\begin{itemize}
    \item $\hat{a}_i^{\text{text}} = f(q_i)$: the model's prediction given only the textual question,
    \item $\hat{a}_i^{\text{speech}} = f(q_i, s_i)$: the model's prediction given both the question and the speech context.
\end{itemize}

We first identify the subset of samples where the model answers the textual question correctly as
\begin{equation}
    \mathcal{D}_{\text{correct}} = \left\{ i \mid \hat{a}_i^{\text{text}} = a_i^{\text{text}} \right\},
\end{equation}
and then compute SCF as the proportion of these samples for which the model faithfully follows the speech context:
\begin{equation}
    \text{SCF} =
    \frac{
        \left|
        \left\{
        i \in \mathcal{D}_{\text{correct}}
        \mid
        \hat{a}_i^{\text{speech}} = a_i^{\text{speech}}
        \right\}
        \right|
    }{
        |\mathcal{D}_{\text{correct}}|
    }.
\end{equation}

A higher SCF indicates that the model is more faithful to the provided speech context rather than relying on its parametric knowledge.

\begin{table}[t!]
\centering
\footnotesize
\setlength{\tabcolsep}{3.5pt}
\renewcommand{\arraystretch}{1.15}
\begin{tabular}{
    >{\raggedright\arraybackslash}p{0.30\columnwidth}
    |
    >{\raggedright\arraybackslash}p{0.62\columnwidth}
}
\toprule\hline
\cellcolor{softgray}\textbf{Field} 
& \cellcolor{softgray}\textbf{Content} \\
\hline

\cellcolor{softgray}\textbf{Speech Context} 
& K/DA는 비디오 게임이던 ``리그 오브 레전드''의 챔피언 아리, 이블린, 카이사, 아칼리으로 구성된 가상의 여성 걸 그룹이다. K/DA라는 이름은 Kill/Death/Assist에서 영어의 한자 씩의 이름으로 왔으며, 팬덤 이름은 블레이즈이다. \newline
\textit{(K/DA is a virtual female group of LoL champions Ahri, Evelynn, Kai'Sa, and Akali. The name derives from Kill/Death/Assist, and the fandom name is Blaze.)} \\
\hline

\cellcolor{softgray}\textbf{Question} 
& 팬덤 이름이 블레이즈인 그룹은? \newline
\textit{(Which group's fandom name is Blaze?)} \\
\hline

\cellcolor{softgray}\textbf{Answer} 
& K/DA \\
\hline

\cellcolor{softgray}
\textbf{Model Response}\newline
\textit{w/ question-only}
& ITZY \\
\hline

\cellcolor{softgray}
\textbf{Model Response}\newline
\textit{w/ speech}\newline
\textit{+ question}
& ITZY \\
\hline

\bottomrule
\end{tabular}
\caption{
A failure case where Qwen3-Omni-30B-A3B-Instruct does not leverage the provided speech context. 
Although the speech context states that the fandom name ``Blaze'' corresponds to K/DA, the model outputs the same incorrect answer, ``ITZY,'' in both the question-only and speech-conditioned settings.
}
\label{tab:case_study_speech_context}
\end{table}

% ============================================================
% CER formula
% ============================================================

\subsection{Evidence Attention Ratio (EAR)}
\label{app:attention_analysis}

To analyze whether models attend to the relevant speech evidence, we measure the evidence attention ratio (EAR), which quantifies how strongly the model focuses on the ground-truth evidence span relative to what would be expected under a uniform attention distribution.

Let $E$ denote the set of tokens belonging to the ground-truth evidence span, $N$ denote the set of non-evidence tokens within the speech context (i.e., $N = S - E$, where $S$ is the full set of speech tokens), and $a_i$ denote the attention weight assigned to token $i$ from a given query token. We first compute the raw attention ratio as
\begin{equation}
\text{EAR}_{\text{raw}} = \frac{\sum_{i \in E} a_i}{\sum_{i \in E} a_i + \sum_{j \in N} a_j}.
\end{equation}
Under a uniform attention distribution, the expected value of $\text{EAR}_{\text{raw}}$ equals the proportion of evidence tokens:
\begin{equation}
\text{EAR}_{\text{uniform}} = \frac{|E|}{|E| + |N|}.
\end{equation}
We define EAR as the ratio of the raw attention ratio to this uniform baseline:
\begin{equation}
\text{EAR} = \frac{\text{EAR}_{\text{raw}}}{\text{EAR}_{\text{uniform}}}.
\end{equation}
An EAR value of $1.0$ indicates that the model distributes attention uniformly across speech tokens, while values greater than $1.0$ indicate above-uniform attention to the evidence span. This normalization ensures that EAR is comparable across samples with different evidence lengths.

For each example, we employ teacher-forced generation: the model-generated answer tokens are appended to the
prompt, and a single forward pass is performed with standard attention to capture the full attention weight matrix. EAR is then computed from the attention of the answer token (e.g., ``(B)'') to all speech tokens, averaged across attention heads. To compare evidence and non-evidence regions, we apply the same formula to each sentence individually, treating its tokens as $E$ and all remaining speech tokens as $N$. We then report EAR for the evidence sentence and the average EAR across all non-evidence sentences.

\section{Additional Results and Analysis}

This section provides supplementary analyses for KoALa-Bench, covering an EAR qualitative example, a speech-context failure case, additional speech translation metrics, and best prompt results across four prompt variants.

\begin{table}[t!]
\centering
\begin{adjustbox}{max width=\columnwidth}
\renewcommand{\arraystretch}{1.5}
\begin{tabular}{l|cc|cc}
\toprule\hline
\multirow{2}{*}{\textbf{Dataset}} & \multicolumn{2}{c|}{\cellcolor{softgray}\textbf{ETRI-TST-COMMON}} & \multicolumn{2}{c}{\cellcolor{softgray}\textbf{ETRI-TST-HE}} \\
& BLEU & METEOR & BLEU & METEOR \\
\hline
Qwen3-omni    & \textbf{28.53} & \textbf{52.02} & \textbf{31.71} & \textbf{55.79} \\
Gemma-3n      & 5.17 & 16.97 & 5.24 & 17.50 \\
GPT-audio     & 26.06 & 49.37 & 28.98 & 52.96 \\
Voxtral       & 23.07 & 45.95 & 26.08 & 50.16 \\
Gemini-flash  & 22.88 & 47.15 & 26.95 & 52.59 \\
\hline
\bottomrule
\end{tabular}
\end{adjustbox}
\caption{Translation performance on ETRI evaluated using BLEU and METEOR scores. Bold indicates the best result.}
\label{tab:speech_translation_bleu_meteor}
\end{table}

\subsection{Failure to Leverage Speech Context}
\label{app:failure_speech_context}
Table~\ref{tab:case_study_speech_context} shows a qualitative failure case where Qwen3-Omni-30B-A3B-Instruct does not leverage the provided speech context. Although the speech input explicitly states that the fandom name ``Blaze'' corresponds to K/DA, the model outputs the same incorrect answer, ``ITZY,'' in both the question-only and speech-conditioned settings. This example illustrates that the model can rely on an unsupported prior association rather than grounding its prediction in the answer-relevant evidence contained in the speech input.

\begin{table*}[t]
\centering
\begin{adjustbox}{max width=\textwidth}
\begin{tabular}{l|l|c|@{\hskip 4pt}c@{\hskip 4pt}c@{\hskip 4pt}c@{\hskip 4pt}c@{\hskip 4pt}c@{\hskip 4pt}c}
\toprule\hline
\multicolumn{2}{l|}{\textbf{Dataset}} & \textbf{Metric} & \textbf{Qwen3-omni} & \textbf{Gemma-3n} & \textbf{GPT-audio} & \textbf{Voxtral} & \textbf{Gemini-flash} & \textbf{Raon} \\
\hline\hline
\rowcolor{softgray}
\multicolumn{9}{c}{\textbf{ASR}} \\
\hline\hline
\multicolumn{2}{l|}{Zeroth} & \multirow{4}{*}{CER $\downarrow$} & \textbf{3.33} / \textbf{\textcolor{blue}{3.91}} & 100$\uparrow$ / \textcolor{blue}{100$\uparrow$} & 6.87 / \textcolor{blue}{9.00} & 40.92 / \textcolor{blue}{39.07} & 13.60 / \textcolor{blue}{14.56} & 4.09 / \textcolor{blue}{5.04} \\
\multicolumn{2}{l|}{Common Voice} & & 4.96 / \textbf{\textcolor{blue}{6.78}} & 100$\uparrow$ / \textcolor{blue}{100$\uparrow$} & 33.05 / \textcolor{blue}{36.21} & 60.10 / \textcolor{blue}{58.98} & 13.74 / \textcolor{blue}{26.74} & \textbf{4.74} / \textcolor{blue}{7.04} \\
\multicolumn{2}{l|}{KsponSpeech-Clean} & & 8.46 & 100$\uparrow$  & 100$\uparrow$  & 62.62  & 83.19 & \textbf{7.55} \\
\multicolumn{2}{l|}{KsponSpeech-Other} & & 7.91 & 100$\uparrow$  & 63.64 & 56.04  & 45.14 & \textbf{7.53} \\
\hline\hline
\rowcolor{softgray}
\multicolumn{9}{c}{\textbf{ST}} \\
\hline\hline
\multicolumn{2}{l|}{ETRI-TST-COMMON} & \multirow{2}{*}{BERTScore $\uparrow$} & \textbf{93.40} & 87.39 & 93.10 & 92.73 & 91.60 & 92.74 \\
\multicolumn{2}{l|}{ETRI-TST-HE} & & \textbf{93.96} & 87.79 & 93.69 & 93.09 & 92.17 & 93.32 \\
\hline\hline
\rowcolor{softgray}
\multicolumn{9}{c}{\textbf{SQA}} \\
\hline\hline
\multirow{2}{*}{Short-Form} & CLIcK & \multirow{3}{*}{Accuracy $\uparrow$} & 64.04 / \textcolor{blue}{62.30} & 35.79 / \textcolor{blue}{35.71} & 61.64 / \textcolor{blue}{60.07} & 42.58 / \textcolor{blue}{42.92} & \textbf{67.27} / \textbf{\textcolor{blue}{66.69}} & 57.42 / \textcolor{blue}{56.59} \\
& KoBEST BoolQ & & 51.34 / \textcolor{blue}{51.16} & 50.89 / \textcolor{blue}{50.54} & 51.88 / \textcolor{blue}{50.45} & 50.54 / \textcolor{blue}{50.54} & \textbf{52.86} / \textbf{\textcolor{blue}{54.92}} & 50.54 / \textcolor{blue}{50.63} \\
\cline{1-2}\cline{4-9}
Long-Form & KCSAT & & \textbf{83.53} / \textbf{\textcolor{blue}{84.71}} & 34.12 / \textcolor{blue}{40.00} & 52.90 / \textcolor{blue}{47.10} & 69.41 / \textcolor{blue}{72.94} & 81.18 / \textcolor{blue}{78.82} & 55.29 / \textcolor{blue}{51.76} \\
\hline\hline
\rowcolor{softgray}
\multicolumn{9}{c}{\textbf{SIF}} \\
\hline\hline
\multicolumn{2}{l|}{KUDGE} & \multirow{4}{*}{\makecell{Score \\ (GPT as Judge)} $\uparrow$} & 71.87 / \textcolor{blue}{71.82} & 71.38 / \textcolor{blue}{70.69} & \textbf{74.07} / \textbf{\textcolor{blue}{73.99}} & 61.97 / \textcolor{blue}{61.70} & 70.29 / \textcolor{blue}{70.39} & 69.07 / \textcolor{blue}{67.13} \\
\multicolumn{2}{l|}{Vicuna} & & 79.64 / \textcolor{blue}{78.43} & 80.21 / \textcolor{blue}{80.00} & \textbf{82.14} / \textbf{\textcolor{blue}{81.79}} & 67.79 / \textcolor{blue}{69.50} & 76.43 / \textcolor{blue}{73.29} & 76.72 / \textcolor{blue}{75.08} \\
\multicolumn{2}{l|}{OpenHermes} &  & 86.54 / \textcolor{blue}{85.19} & 84.62 / \textcolor{blue}{85.96} & \textbf{89.42} / \textbf{\textcolor{blue}{89.62}} & 69.10 / \textcolor{blue}{69.62} & 82.69 / \textcolor{blue}{80.71} & 79.74 / \textcolor{blue}{78.72} \\
\multicolumn{2}{l|}{Alpaca} & & 84.06 / \textcolor{blue}{83.04} & 82.97 / \textcolor{blue}{83.36} & \textbf{90.58} / \textbf{\textcolor{blue}{90.58}} & 72.90 / \textcolor{blue}{72.46} & 85.94 / \textcolor{blue}{86.88} & 78.26 / \textcolor{blue}{77.68} \\
\hline
\bottomrule
\end{tabular}
\end{adjustbox}
\caption{
Best prompt performance across ASR, ST, SQA, and SIF. 
For each model and task, we evaluate four prompt variants and report only the best score among them. 
ASR, SQA, and SIF are evaluated on clean and noisy audio when applicable. 
KsponSpeech-Clean/Other denote different subsets. 
$100\uparrow$ indicates CER $>$ 100\%. 
Blue indicates noisy results.
}
\label{tab:all_results_best_prompt}
\end{table*}

% ========== Specific bench mark: SCA-QA & PA-QA
\begin{table*}[t!]
\centering
\begin{adjustbox}{max width=\textwidth}
\begin{tabular}{l|l|c|cccccc}
\toprule\hline
\multicolumn{2}{l|}{\textbf{Dataset}} & \textbf{Metric} & \textbf{Qwen3-omni} & \textbf{Gemma-3n} & \textbf{GPT-audio} & \textbf{Voxtral} & \textbf{Gemini-flash} & \textbf{Raon} \\
\hline\hline
\multicolumn{9}{c}{\cellcolor{softgray}\textbf{SCA-QA}}\\
\hline\hline
\multicolumn{2}{l|}{K-history (Before)}   & \multirow{4}{*}{\makecell{Text-only \\ Accuracy} $\uparrow$} & 55.45  & 48.51  & 69.30  & 41.58 & \textbf{79.21} & 52.48 \\
\multicolumn{2}{l|}{K-history (After Chosun)}    & & 79.27 & 71.95  & 79.30 & 51.22 & \textbf{87.80} & 65.85 \\
\multicolumn{2}{l|}{K-sports}             & & 55.68 & 53.41 & 69.30  & 27.27  & \textbf{78.41} & 55.68 \\
\multicolumn{2}{l|}{K-pop}                & & 67.96 & 68.93 & 69.90 & 42.72  & \textbf{89.32} & 55.34 \\
\hline
\multicolumn{2}{l|}{K-history (Before)}   &  \multirow{4}{*}{SCF Score $\uparrow$} & \textbf{94.64} & 67.35 & 61.40 & 85.71 & 66.25 & 38.78 \\
\multicolumn{2}{l|}{K-history (After Chosun)}    & & \textbf{92.31} & 45.76 & 32.30 & 85.71 & 59.72 & 23.53 \\
\multicolumn{2}{l|}{K-sports}             & & 93.88 & 76.60 & 39.30 & \textbf{95.83} & 86.96 & 28.57 \\
\multicolumn{2}{l|}{K-pop}                & & \textbf{95.71} & 73.24 & 37.50 & 88.64 & 78.26 & 33.33 \\
\hline\hline
\multicolumn{9}{c}{\cellcolor{softgray}\textbf{PA-QA}}\\
\hline\hline
\multicolumn{2}{l|}{MC-Test Total} & \multirow{5}{*}{Accuracy $\uparrow$} & \textbf{93.54} / \textbf{\textcolor{blue}{93.85}} & 48.92 / \textcolor{blue}{48.92} & 77.23 / \textcolor{blue}{79.69} & 84.92 / \textcolor{blue}{86.15} & 92.31 / \textcolor{blue}{92.00} & 40.98 / \textcolor{blue}{41.28} \\
\cline{1-2}\cline{4-9}
\multirow{4}{*}{\makecell{Position\\- Aware}} & Front &  & 91.93 / \textcolor{blue}{91.93} & 45.34 / \textcolor{blue}{44.72} & 72.05 / \textcolor{blue}{75.78} & 86.34 / \textcolor{blue}{85.09} & \textbf{95.03} / \textbf{\textcolor{blue}{93.17}} & 43.48 / \textcolor{blue}{42.03} \\
& Front-Middle  & & \textbf{96.70} / \textbf{\textcolor{blue}{94.51}} & 51.65 / \textcolor{blue}{53.85} & 83.52 / \textcolor{blue}{83.52} & 82.42 / \textcolor{blue}{85.71} & 89.01 / \textcolor{blue}{87.91} & 42.70 / \textcolor{blue}{44.94} \\
& Middle-Last  & & \textbf{87.80} / \textbf{\textcolor{blue}{87.80}} & 47.56 / \textcolor{blue}{47.56} & 74.39 / \textcolor{blue}{70.73} & 76.83 / \textcolor{blue}{81.71} & 82.93 / \textcolor{blue}{85.37} & 32.47 / \textcolor{blue}{32.47} \\
& Last  & & \textbf{94.12} / \textcolor{blue}{98.04} & 45.10 / \textcolor{blue}{45.10} & 80.39 / \textcolor{blue}{86.27} & 82.35 / \textcolor{blue}{82.35} & 92.16 / \textbf{\textcolor{blue}{98.04}} & 40.00 / \textcolor{blue}{38.18} \\
\hline
\bottomrule
\end{tabular}
\end{adjustbox}
\caption{
Best prompt performance on speech faithfulness tasks. 
For each model and task, we evaluate four prompt variants and report only the best score among them. 
SCA-QA reports text-only accuracy and SCF score, while PA-QA reports accuracy by evidence position. 
Blue indicates noisy results.
}
\label{tab:faithfulness_results_best_prompt}
\end{table*}

\subsection{Additional Metrics for Speech Translation.}

We additionally provide translation evaluation results using BLEU and METEOR.
As shown in Table~\ref{tab:speech_translation_bleu_meteor}, Qwen3-Omni achieved the best performance on both ETRI-COMMON and ETRI-HE, obtaining the highest BLEU and METEOR across the compared models.

\begin{figure*}[h]
\centering
\begin{tcolorbox}[
  width=\textwidth,
  title={\textbf{EAR Example Instance}\\\textit{\small PA-QA sample \texttt{000541} used in
Figure~\ref{fig:ear_comparison}.}},
  colback=gray!5,
  colframe=tealframe,
  colbacktitle=tealtitle,
  coltitle=black,
  fonttitle=\large,
  boxrule=0.5pt,
  top=2pt, bottom=2pt, left=4pt, right=4pt,
]
{\footnotesize\ttfamily
\textbf{Context:}\\
농구를 하는 남자 넷이 있었습니다. 그들은 야구, 미식축구,
축구는 하지 않았습니다. 그들의 이름은 세스, 태너, 헨리,
라이언이었습니다. 그들 중 한 명은 서부 지역에서 최고의
슈팅 실력을 자랑했습니다. 그는 너무 훌륭해서 거의 슛을
놓치지 않았습니다. 세상 모든 사람들은 그처럼 훌륭해지기를
원했습니다. 태너는 거의 슛을 놓치지 않는 사람이었습니다.
그는 매일 농구를 했습니다. 그는 슛을 던지고, 드리블하고,
매우 빠르게 달릴 수 있었습니다. 하지만 덩크슛은 할 수
없었습니다. 그는 농구를 너무 잘해서 훗스터스, 슈터스,
드리블러스, 덩커스 같은 팀들이 그를 영입하려고 했습니다.
그는 팀을 선택하는 데 매우 어려움을 겪었습니다.
그는 빠르게 팀을 선택해야 했습니다. 그들은 선수가
필요했기 때문에 태너는 팀을 선택해야 했습니다.
그는 훗스터스를 선택했습니다. 그들은 그의 가장 친한
친구들이었습니다. 태너는 그들과 함께 많은 경기를 치렀고,
심지어 그들의 에이스 선수였습니다. 그는 농구를 정말
즐겼고, 경기를 하는 것이 매우 즐거웠습니다. 그는 너무
즐거워서 오랫동안 농구를 했습니다.
\\
\\
\textbf{Prompt\_ko:} 태너는 무엇을 할 수 없었을까요?
\\
\\
\textbf{Choices\_ko:}\\
(A) 빠르게 달린다 (B) 덩크슛을 한다
(C) 슛을 한다 (D) 드리블한다
\\
\\
\textbf{Ground truth:} (B)
\\
\\
\textbf{Evidence:} 하지만 덩크슛은 할 수 없었습니다.
(position: 47\%, sentence 10/19)
\\
\\
\textbf{Non-evidence (avg):} 나머지 18개 문장의 평균
}
\end{tcolorbox}
\caption{
Representative PA-QA instance used for EAR analysis. 
The annotated evidence sentence directly supports the correct answer, while the remaining sentences are treated as non-evidence for comparison.
}
\label{fig:ear_example}
\end{figure*}

\begin{figure*}[h]
\centering
\begin{tcolorbox}[
  width=\textwidth,
  title={\textbf{EAR Example Instance (English translation)}\\\textit{\small English rendering of the Korean PA-QA sample \texttt{000541} shown in Figure~\ref{fig:ear_example}.}},
  colback=gray!5,
  colframe=tealframe,
  colbacktitle=tealtitle,
  coltitle=black,
  fonttitle=\large,
  boxrule=0.5pt,
  top=2pt, bottom=2pt, left=4pt, right=4pt,
]

{\footnotesize\itshape
\textbf{Context (English translation):}\\
There were four boys who played basketball. They did not play baseball, football, or soccer. Their names were Seth, Tanner, Henry, and Ryan. One of them had the best shooting skills in the west. He was so good that he almost never missed a shot. Everyone in the world wanted to be as good as him. Tanner was the one who almost never missed a shot. He played basketball every day. He could shoot, dribble, and run very fast. However, he could not dunk. He was so good at basketball that teams such as the Hoopsters, the Shooters, the Dribblers, and the Dunkers tried to recruit him. He had great difficulty choosing a team. He had to choose one quickly, because the teams needed a player. He chose the Hoopsters. They were his best friends. Tanner played many games with them and was even their star player. He truly enjoyed basketball and had a lot of fun playing. He enjoyed it so much that he played basketball for a long time.
\\
\\
\textbf{Prompt:} What was Tanner unable to do?
\\
\\
\textbf{Choices:}\\
(A) run fast (B) dunk (C) shoot (D) dribble
\\
\\
\textbf{Ground truth:} (B)
\\
\\
\textbf{Evidence:} However, he could not dunk. (position: 47\%, sentence 10/19)
\\
\\
\textbf{Non-evidence (avg):} average over the remaining 18 sentences
}
\end{tcolorbox}
\caption{English translation of the PA-QA sample shown in Figure~\ref{fig:ear_example}. The translation is provided for readability only; the Korean version is the one used in our experiments.}
\label{fig:ear_example_en}
\end{figure*}

\subsection{Best Prompt Reporting across Four Variants}
\label{sec:best_prompt_reporting_four_variants}
% For each task, we evaluate each model with four prompt variants and report the best score among them. 
% Tables~\ref{tab:all_results} and~\ref{tab:faithfulness_results} summarize the resulting performance across the general and QA-based evaluation tasks.

For each task, we evaluate each model using four semantically equivalent prompt variants and report the best score among them. 
Therefore, Tables~\ref{tab:all_results_best_prompt} and~\ref{tab:faithfulness_results_best_prompt} should be interpreted as best prompt result tables, rather than prompt sensitivity or prompt average tables. 
This reporting protocol reduces the effect of incidental prompt wording while keeping the task objective and answer format fixed across variants.

Table~\ref{tab:all_results_best_prompt} summarizes the best prompt performance on the general evaluation tasks, including ASR, ST, SQA, and SIF. 
Overall, Qwen3-omni achieves the strongest performance on Korean ASR and ST, while GPT-audio obtains the highest scores on most SIF datasets. 
For SQA, Gemini-flash performs best on short-form QA datasets, whereas Qwen3-omni shows stronger performance on the long-form KCSAT subset. 

Table~\ref{tab:faithfulness_results_best_prompt} summarizes the best prompt performance on the faithfulness-oriented evaluation tasks, including SCA-QA and PA-QA. 
In SCA-QA, Gemini-flash achieves the highest text-only accuracy, but Qwen3-omni shows stronger SCF scores in most domains, indicating better adherence to the provided speech context when it conflicts with prior knowledge. 
In PA-QA, Qwen3-omni achieves the best overall accuracy and remains competitive across evidence positions, suggesting strong long-context speech understanding under the position-aware evaluation setting.

\subsection{EAR Qualitative Analysis Example}
\label{sec:ear_example}

Figure~\ref{fig:ear_example} shows a representative PA-QA instance.
The example consists of a Korean speech context, a multiple-choice question, the ground-truth answer, and an annotated evidence sentence with its relative position in the context. 
The evidence sentence directly supports the correct option, while the remaining sentences are treated as non-evidence, enabling us to compare model attribution to answer-relevant information against attribution to irrelevant context.
An English translation of this sample is provided in Figure~\ref{fig:ear_example_en} for reference.

\section{Lists of Prompts}
This appendix provides the full set of prompts used in KoALa-Bench. 
We include the evaluation prompts for all benchmark tasks, the LLM as judge prompt used for automatic response quality assessment, and the prompts used in the SCA-QA construction pipeline. 
These prompt lists are provided to make the evaluation and dataset construction procedures transparent and reproducible.
All prompts are written in Korean, since KoALa-Bench targets Korean speech understanding. 
For readability, every Korean prompt figure is immediately followed by an English translation of the same figure; the translations are provided for reference only and are never given to the models.

% ============================================================
% Full List of Evaluation Prompts
% ============================================================
\begin{figure*}[!t]
\centering
\begin{tcolorbox}[
  width=\textwidth,
  title={\textbf{Prompts for LLM as Judge.}},
  colback=white,
  colframe=steelblue!70!black,
  colbacktitle=steelblue!30,
  coltitle=black,
  fonttitle=\large\bfseries,
  boxrule=0.5pt,
]

\textbf{SIF (Speech Instruction Following)}\\
\textit{Objective: Evaluate whether the model's response correctly follows the given instruction.}

\begin{tcolorbox}[
  width=\linewidth,
  colback=gray!5, colframe=gray!50, boxrule=0.3pt, top=2pt, bottom=2pt,
]
\textbf{Prompts:}\\
{\small\ttfamily
다음은 질문과 두 개의 답변입니다. 생성된 답변을 참조 답변과 비교하여 평가해 주세요.\\

질문: \{question\} \\

참조 답변 (GT): \{reference\_answer\} \\

생성된 답변: \{generated\_answer\} \\

평가 기준:\\
1. 정확성: 생성된 답변이 질문에 올바르게 답하고 있는가?\\
2. 완전성: 생성된 답변이 참조 답변만큼 완전한 정보를 제공하는가?\\
3. 관련성: 생성된 답변이 질문과 관련이 있는가?\\

0.0부터 1.0 사이의 점수를 부여해 주세요. 점수만 숫자로 응답해 주세요.
}
\end{tcolorbox}
\end{tcolorbox}
\caption{
Prompt template for LLM based judging in SIF evaluation. 
Given a question, a reference answer, and a generated answer, the judge outputs a scalar quality score based on accuracy, completeness, and relevance.
}
\label{fig:llm_judge_prompt}
\end{figure*}

\begin{figure*}[!t]
\centering
\begin{tcolorbox}[
  width=\textwidth,
  title={\textbf{Prompts for LLM as Judge (English translation)}},
  colback=white,
  colframe=steelblue!70!black,
  colbacktitle=steelblue!30,
  coltitle=black,
  fonttitle=\large\bfseries,
  boxrule=0.5pt,
]

\textbf{SIF (Speech Instruction Following)}\\
\textit{Objective: Evaluate whether the model's response correctly follows the given instruction.}

\begin{tcolorbox}[
  width=\linewidth,
  colback=gray!5, colframe=gray!50, boxrule=0.3pt, top=2pt, bottom=2pt,
]

{\small\itshape
Below are a question and two answers. Please evaluate the generated answer by comparing it with the reference answer.\\

Question: \{question\} \\

Reference answer (GT): \{reference\_answer\} \\

Generated answer: \{generated\_answer\} \\

Evaluation criteria:\\
1. Accuracy: does the generated answer correctly answer the question?\\
2. Completeness: does the generated answer provide information as complete as the reference answer?\\
3. Relevance: is the generated answer relevant to the question?\\

Please assign a score between 0.0 and 1.0. Respond with the score only, as a number.
}
\end{tcolorbox}
\end{tcolorbox}
\caption{English translation of the prompt shown in Figure~\ref{fig:llm_judge_prompt}. The translation is provided for readability only; the Korean version is the one used in our experiments.}
\label{fig:llm_judge_prompt_en}
\end{figure*}

\begin{figure*}[!t]
\centering
\begin{tcolorbox}[
  width=\textwidth,
  title={\textbf{Prompts for KoALa-Bench Evaluation Tasks (Part I: ASR, ST, and SIF)}},
  colback=white,
  colframe=steelblue!70!black,
  colbacktitle=steelblue!30,
  coltitle=black,
  fonttitle=\large\bfseries,
  boxrule=0.5pt,
]

%% ASR %%
\textbf{ASR (Automatic Speech Recognition)}\\
\textit{Objective: Transcribe the given speech into Korean text.}

\begin{tcolorbox}[
  width=\linewidth,
  colback=gray!5,
  colframe=gray!50,
  boxrule=0.3pt,
  top=2pt,
  bottom=2pt,
]
\textbf{Prompts:}\\
{\small\ttfamily
1) 주어진 음성을 한국어로 전사하고 전사된 한국어 텍스트만 출력해 주세요.\\
2) 다음 음성을 한국어로 전사해 주시고 설명이나 추가 문장 없이 전사 결과만 제공해 주세요.\\
3) 제공된 음성의 내용을 한국어로 정확하게 전사해 주시고 전사된 문장만 출력해 주세요.\\
4) 맥락으로 주어진 음성을 한국어로 전사해 주시되 다른 설명 없이 전사된 한국어 텍스트만 응답해 주세요.
}
\end{tcolorbox}

\vspace{4pt}

%% ST %%
\textbf{ST (Speech Translation)}\\
\textit{Objective: Translate the given speech into Korean.}

\begin{tcolorbox}[
  width=\linewidth,
  colback=gray!5,
  colframe=gray!50,
  boxrule=0.3pt,
  top=2pt,
  bottom=2pt,
]
\textbf{Prompts:}\\
{\small\ttfamily
1) 이 오디오를 한국어로 번역해 주세요. 번역 결과만 출력하고 다른 텍스트는 출력하지 마세요.\\
2) 다음 문장을 의미를 변경하지 말고 자연스러운 한국어로 번역하시오.\\
3) 다음 영어 문장을 정보의 추가나 삭제 없이 정확하게 한국어로 번역하시오.\\
4) 다음 문장의 의미를 그대로 유지한 채 한국어로 번역하고 번역 결과만 출력하시오.
}
\end{tcolorbox}

\vspace{4pt}

%% SIF %%
\textbf{SIF (Speech Instruction Following)}\\
\textit{Objective: Assess the model's ability to follow spoken instructions.}

\begin{tcolorbox}[
  width=\linewidth,
  colback=gray!5,
  colframe=gray!50,
  boxrule=0.3pt,
  top=2pt,
  bottom=2pt,
]
\textbf{Prompts:}\\
{\small\ttfamily
1) 음성에서 제시된 질문을 참고해 주시기 바랍니다.\\
2) 음성의 정보를 활용하여 다음 질문에 답해 주세요.\\
3) 음성을 듣고 다음 질문에 답해 주세요.\\
4) 음성 내용을 바탕으로 다음 질문에 답해 주세요.
}
\end{tcolorbox}

\end{tcolorbox}
\caption{
Prompt templates used for general speech evaluation tasks in KoALa-Bench. 
Part I includes ASR, speech translation, and speech instruction following tasks. 
Each task is evaluated with four prompt variants while maintaining the same task objective.
}
\label{fig:eval_prompts_part1}
\end{figure*}

\begin{figure*}[!t]
\centering
\begin{tcolorbox}[
  width=\textwidth,
  title={\textbf{Prompts for KoALa-Bench Evaluation Tasks, Part I (English translation)}},
  colback=white,
  colframe=steelblue!70!black,
  colbacktitle=steelblue!30,
  coltitle=black,
  fonttitle=\large\bfseries,
  boxrule=0.5pt,
]

%% ASR %%
\textbf{ASR (Automatic Speech Recognition)}\\
\textit{Objective: Transcribe the given speech into Korean text.}

\begin{tcolorbox}[
  width=\linewidth,
  colback=gray!5,
  colframe=gray!50,
  boxrule=0.3pt,
  top=2pt,
  bottom=2pt,
]

{\small\itshape
1) Transcribe the given speech into Korean and output only the transcribed Korean text.\\
2) Transcribe the following speech into Korean and provide only the transcription, without explanations or additional sentences.\\
3) Accurately transcribe the content of the provided speech into Korean and output only the transcribed sentences.\\
4) Transcribe the speech given as context into Korean, and respond with only the transcribed Korean text without any other explanation.
}
\end{tcolorbox}

\vspace{4pt}

%% ST %%
\textbf{ST (Speech Translation)}\\
\textit{Objective: Translate the given speech into Korean.}

\begin{tcolorbox}[
  width=\linewidth,
  colback=gray!5,
  colframe=gray!50,
  boxrule=0.3pt,
  top=2pt,
  bottom=2pt,
]

{\small\itshape
1) Translate this audio into Korean. Output only the translation and no other text.\\
2) Translate the following sentence into natural Korean without changing its meaning.\\
3) Translate the following English sentence into Korean accurately, without adding or omitting information.\\
4) Translate the following sentence into Korean while preserving its meaning exactly, and output only the translation.
}
\end{tcolorbox}

\vspace{4pt}

%% SIF %%
\textbf{SIF (Speech Instruction Following)}\\
\textit{Objective: Assess the model's ability to follow spoken instructions.}

\begin{tcolorbox}[
  width=\linewidth,
  colback=gray!5,
  colframe=gray!50,
  boxrule=0.3pt,
  top=2pt,
  bottom=2pt,
]

{\small\itshape
1) Please refer to the question presented in the speech.\\
2) Using the information in the speech, please answer the following question.\\
3) Listen to the speech and answer the following question.\\
4) Based on the content of the speech, answer the following question.
}
\end{tcolorbox}

\end{tcolorbox}
\caption{English translation of the prompt shown in Figure~\ref{fig:eval_prompts_part1}. The translation is provided for readability only; the Korean version is the one used in our experiments.}
\label{fig:eval_prompts_part1_en}
\end{figure*}

\begin{figure*}[t!]
\centering
\begin{tcolorbox}[
  width=\textwidth,
  title={\textbf{Prompts for KoALa-Bench Evaluation Tasks (Part II: SQA, PA-QA, and SCA-QA)}},
  colback=white,
  colframe=steelblue!70!black,
  colbacktitle=steelblue!30,
  coltitle=black,
  fonttitle=\large\bfseries,
  boxrule=0.5pt,
]

%% SQA / PA-QA %%
\textbf{SQA / PA-QA}\\
\textit{Objective: Answer questions based on the given speech.}

\begin{tcolorbox}[
  width=\linewidth,
  colback=gray!5,
  colframe=gray!50,
  boxrule=0.3pt,
  top=2pt,
  bottom=2pt,
]
\textbf{Prompt Prefix:}\\
{\small\ttfamily
1) <One-shot prompt> 다음 음성을 듣고 질문에 맞는 답을 고르세요.\\
2) <One-shot prompt> 주어진 음성을 듣고 가장 적절한 답을 선택하세요.\\
3) <One-shot prompt> 음성을 잘 듣고 알맞은 답을 골라 주세요.\\
4) <One-shot prompt> 아래 음성의 내용을 참고하여 올바른 답을 선택하세요.}

\vspace{3pt}
\textbf{Answer Suffix:} {\small\ttfamily \textbackslash n답:}

\vspace{3pt}
\textbf{One-shot prompt:}\\
{\small\ttfamily
예시:\textbackslash n질문: 철수는 어디에서 친구를 만났나요?\textbackslash n(A) 학교 (B) 공원 (C) 도서관 (D) 카페\textbackslash n답: (B)\textbackslash n\textbackslash n
}
\end{tcolorbox}

\vspace{4pt}

%% SCA-QA %%
\textbf{SCA-QA (Speech Context Aware Question Answering)}\\
\textit{Objective: Answer questions based solely on the text question, comparing speech-grounded responses.\\ A one-shot example is used to elicit answers in a consistent, parseable format.}

\begin{tcolorbox}[
  width=\linewidth,
  colback=gray!5,
  colframe=gray!50,
  boxrule=0.3pt,
  top=2pt,
  bottom=2pt,
]
\textbf{Prompt (Text-only):}\\
{\small\ttfamily
1) <One-shot example> 다음 질문에 답변하세요.\\
2) <One-shot example> 질문에 맞는 답을 고르세요.\\
3) <One-shot example> 아래 질문에 가장 적절한 답을 선택하세요.\\
4) <One-shot example> 주어진 선택지 중 올바른 답을 고르세요.\\
}

\vspace{4pt}
\textbf{Prompt (with speech context):}\\
{\small\ttfamily
1) <One-shot prompt> 음성에서 들려준 내용만을 근거로 다음 질문에 답변하세요.\\
2) <One-shot prompt> 음성에서 제공된 정보에만 기반하여 질문에 맞는 답을 고르세요. \\
3) <One-shot prompt> 반드시 음성에서 들은 내용을 바탕으로 아래 질문에 가장 적절한 \\답을 선택하세요.\\
4) <One-shot prompt> 음성의 내용만을 참고하여 주어진 선택지 중 올바른 답을 고르세요.\\
}

\textbf{One-shot prompt:}\\
{\small\ttfamily
예시:\\질문: 철수는 어디에서 친구를 만났나요?\\(A) 학교 (B) 공원 (C) 도서관 (D) 카페\\답: (B)
}
\end{tcolorbox}

\end{tcolorbox}
\caption{Prompt templates used for QA-based evaluation tasks in KoALa-Bench. 
Part II includes SQA, PA-QA, and SCA-QA. 
For multiple-choice QA tasks, we use one-shot prompting and a fixed answer suffix to encourage responses in a consistent and parseable format.
}
\label{fig:eval_prompts_part2}
\end{figure*}

\begin{figure*}[t!]
\centering
\begin{tcolorbox}[
  width=\textwidth,
  title={\textbf{Prompts for KoALa-Bench Evaluation Tasks, Part II (English translation)}},
  colback=white,
  colframe=steelblue!70!black,
  colbacktitle=steelblue!30,
  coltitle=black,
  fonttitle=\large\bfseries,
  boxrule=0.5pt,
]

%% SQA / PA-QA %%
\textbf{SQA / PA-QA}\\
\textit{Objective: Answer questions based on the given speech.}

\begin{tcolorbox}[
  width=\linewidth,
  colback=gray!5,
  colframe=gray!50,
  boxrule=0.3pt,
  top=2pt,
  bottom=2pt,
]

{\small\itshape
1) <One-shot prompt> Listen to the following speech and choose the answer that matches the question.\\
2) <One-shot prompt> Listen to the given speech and select the most appropriate answer.\\
3) <One-shot prompt> Listen carefully to the speech and pick the correct answer.\\
4) <One-shot prompt> Referring to the content of the speech below, select the correct answer.

\vspace{3pt}
Answer suffix: ``\textbackslash nAnswer:''

\vspace{3pt}
One-shot prompt: Example:\textbackslash nQuestion: Where did Cheolsu meet his friend?\textbackslash n(A) school (B) park (C) library (D) cafe\textbackslash nAnswer: (B)
}
\end{tcolorbox}

\vspace{4pt}

%% SCA-QA %%
\textbf{SCA-QA (Speech Context Aware Question Answering)}\\
\textit{Objective: Answer questions based solely on the text question, comparing speech-grounded responses.\\ A one-shot example is used to elicit answers in a consistent, parseable format.}

\begin{tcolorbox}[
  width=\linewidth,
  colback=gray!5,
  colframe=gray!50,
  boxrule=0.3pt,
  top=2pt,
  bottom=2pt,
]

{\small\itshape
\textbf{Text-only.}
1) <One-shot example> Answer the following question.\\
2) <One-shot example> Choose the answer that matches the question.\\
3) <One-shot example> Select the most appropriate answer to the question below.\\
4) <One-shot example> Choose the correct answer among the given options.

\vspace{4pt}
\textbf{With speech context.}
1) <One-shot prompt> Answer the following question based only on what was said in the speech.\\
2) <One-shot prompt> Based only on the information provided in the speech, choose the answer that matches the question.\\
3) <One-shot prompt> Relying strictly on what you heard in the speech, select the most appropriate answer to the question below.\\
4) <One-shot prompt> Referring only to the content of the speech, choose the correct answer among the given options.

\vspace{4pt}
\textbf{One-shot prompt.} Example: Question: Where did Cheolsu meet his friend? (A) school (B) park (C) library (D) cafe / Answer: (B)
}
\end{tcolorbox}

\end{tcolorbox}
\caption{English translation of the prompt shown in Figure~\ref{fig:eval_prompts_part2}. The translation is provided for readability only; the Korean version is the one used in our experiments.}
\label{fig:eval_prompts_part2_en}
\end{figure*}

%% Stage 1 %%
\begin{figure*}[t!]
\centering
\begin{tcolorbox}[
  width=\textwidth,
  title={\textbf{Stage 1: Question and Answer Generation}\\\textit{\small Objective: Generate a factual question and a short answer from the given context.}},
  colback=gray!5,
  colframe=tealframe,
  colbacktitle=tealtitle,
  coltitle=black,
  fonttitle=\large,
  boxrule=0.5pt,
  top=2pt, bottom=2pt, left=4pt, right=4pt,
]
{\footnotesize\ttfamily
문단(context)을 읽고, 그 내용으로 답할 수 있는 자세한 질문을 만든 뒤, 정답을 한 단어(또는 짧은 구)로 써 주세요.\\
- 질문: 문단에 나온 사실을 묻는 구체적인 질문. 질문 문장 안에 정답 단어를 넣지 마세요.\\
- 답변: 문단에서 묻는 대상(인물, 사건, 장소, 작품명 등)을 짧게 한 단어로.\\
\\
{[}예시 1{]}\\
context: 살수 대첩은 제2차 고수 전쟁을 고구려의 승리로 장식한 전투이다. 평양을 직공했던 수나라의 30만 군대가 살수를 건너 회군하던 중 을지문덕 군의 공격을 받아 궤멸적인 피해를 입고 패주했다. 귀주 대첩, 한산도 대첩과 함께 한국사 3대 대첩 중 하나로 불린다.\\
질문: 제2차 고수 전쟁을 고구려의 승리로 장식한 전투이며, 수나라 30만 군대가 을지문덕의 공격을 받아 궤멸적 피해를 입고 패주한, 한국사 3대 대첩 중 하나로 불리는 전투는 무엇인가?\\
답변 : 살수 대첩\\
\\
{[}예시 2{]}\\
context: 안시성 전투는 제1차 고당 전쟁 중 당 태종, 이세적이 지휘하는 군대가 대규모로 고구려를 침공하여 2개월 가량 안시성을 포위하고 공격했으나, 끈질긴 항전으로 이를 물리치고 고구려가 승리한 싸움이다.\\
질문: 제1차 고당 전쟁 중 당 태종과 이세적이 이끄는 군대가 2개월 동안 포위·공격했으나 고구려가 끈질긴 항전 끝에 승리한 전투는 무엇인가?\\
답변 : 안시성 전투\\
\\
{[}예시 3{]}\\
context: 황산벌 전투는 660년 8월 20일에서 8월 21일 황산벌에서 신라 군과 백제 군 사이에 일어났던 전투이다. 삼국유사 태종무열왕조, 삼국사기 계백 열전 등에 전투 내용이 나온다.\\
질문: 660년 8월 20일과 21일 사이에 황산벌에서 신라 군과 백제 군 간에 벌어진 전투로, 삼국유사와 삼국사기 계백 열전에 기록된 전투의 이름은 무엇인가?\\
답변 : 황산벌 전투\\
\\
---\\
아래 문단을 보고 같은 형식으로 질문과 답변을 작성하세요. (질문: ... / 답변 : ...)\\
context:\\
\{context\}\\
\\
질문:
}
\end{tcolorbox}
\caption{Prompt used for stage 1 of the SCA-QA dataset construction pipeline. Given a context passage, the model generates a factual question and a short answer grounded in the passage.}
\label{fig:sca_qa_stage1_prompt}
\end{figure*}

\begin{figure*}[t!]
\centering
\begin{tcolorbox}[
  width=\textwidth,
  title={\textbf{Stage 1: Question and Answer Generation --- English translation}\\\textit{\small Objective: Generate a factual question and a short answer from the given context. Shown for readability; the Korean prompt in Figure~\ref{fig:sca_qa_stage1_prompt} is the one used.}},
  colback=gray!5,
  colframe=tealframe,
  colbacktitle=tealtitle,
  coltitle=black,
  fonttitle=\large,
  boxrule=0.5pt,
  top=2pt, bottom=2pt, left=4pt, right=4pt,
]

{\footnotesize\itshape
Read the passage (context), write a detailed question that can be answered from it, and then give the answer as a single word (or a short phrase).\\
- Question: a specific question about a fact stated in the passage. Do not include the answer word in the question itself.\\
- Answer: the target being asked about in the passage (a person, event, place, title, etc.), given briefly as a single word.\\
\\
{[}Example 1{]}\\
context: The Battle of Salsu was the battle that sealed the Second Goguryeo--Sui War as a Goguryeo victory. The 300,000-strong Sui army that had struck directly at Pyongyang was attacked by Eulji Mundeok's forces while crossing the Salsu River on its retreat, suffering devastating losses and fleeing in defeat. Together with the Battle of Gwiju and the Battle of Hansando, it is counted among the three great victories of Korean history.\\
Question: Which battle sealed the Second Goguryeo--Sui War as a Goguryeo victory, in which the 300,000-strong Sui army was attacked by Eulji Mundeok and fled after devastating losses, and is counted among the three great victories of Korean history?\\
Answer: the Battle of Salsu\\
\\
{[}Example 2{]}\\
context: The Battle of Ansi Fortress was a battle of the First Goguryeo--Tang War in which the army commanded by Emperor Taizong of Tang and Li Shiji invaded Goguryeo on a large scale and besieged and attacked Ansi Fortress for about two months, but was repelled by tenacious resistance, resulting in a Goguryeo victory.\\
Question: Which battle of the First Goguryeo--Tang War, in which the army led by Emperor Taizong and Li Shiji besieged and attacked for two months, ended in a Goguryeo victory after tenacious resistance?\\
Answer: the Battle of Ansi Fortress\\
\\
{[}Example 3{]}\\
context: The Battle of Hwangsanbeol was fought between the Silla and Baekje armies at Hwangsanbeol from 20 to 21 August 660. Accounts of the battle appear in the Taejong Muyeol section of the Samguk Yusa and in the biography of Gyebaek in the Samguk Sagi.\\
Question: What is the name of the battle fought between the Silla and Baekje armies at Hwangsanbeol on 20--21 August 660, recorded in the Samguk Yusa and in the biography of Gyebaek in the Samguk Sagi?\\
Answer: the Battle of Hwangsanbeol\\
\\
---\\
Read the passage below and write a question and an answer in the same format. (Question: ... / Answer: ...)\\
context:\\
\{context\}\\
\\
Question:
}
\end{tcolorbox}
\caption{English translation of the prompt shown in Figure~\ref{fig:sca_qa_stage1_prompt}. The translation is provided for readability only; the Korean version is the one used in our experiments.}
\label{fig:sca_qa_stage1_prompt_en}
\end{figure*}

%% Stage 2 %%
\begin{figure*}[t!]
\centering
\begin{tcolorbox}[
  width=\textwidth,
  title={\textbf{Stage 2: Entity Replacement}\\\textit{\small Objective: Find a different entity of the same type to replace the original answer.}},
  colback=gray!5,
  colframe=tealframe,
  colbacktitle=tealtitle,
  coltitle=black,
  fonttitle=\large,
  boxrule=0.5pt,
  top=2pt, bottom=2pt, left=4pt, right=4pt,
]
{\footnotesize\ttfamily
컨텍스트와 현재 엔티티가 주어집니다. 이 엔티티와 **같은 유형**의 **다른** 엔티티 하나만 골라서 알려 주세요. (예: 역사 전투 → 다른 전투명, 걸그룹 → 다른 걸그룹명) 문장을 바꾸지 말고, 새 엔티티 이름만 출력하세요.\\
IMPORTANT!!! 컨텍스트 안에 있는 단어를 새로운 엔티티로 작성하지 마세요. 새로운 엔티티는 문장 안에 없어야 합니다.\\
\\
{[}예시 1{]}\\
문장: 살수 대첩은 제2차 고수 전쟁 을 고구려 의 승리로 장식한 전투이다. 평양을 직공했던 수나라의 30만 군대가 살수를 건너 회군하던 중 을지문덕 군의 공격을 받아 궤멸적인 피해를 입고 패주했다. 귀주 대첩, 한산도 대첩과 함께 한국사 3대 대첩 중 하나로 불린다.\\
현재 엔티티: 살수 대첩\\
→ 새로운 엔티티: 안시성 전투\\
불가능한 엔티티 : 한산도 대첩(컨텍스트 안에 존재함)\\
\\
{[}예시 2{]}\\
문장: 크레용팝은 대한민국의 오인조 걸 그룹이다...\\
현재 엔티티: 크레용팝\\
→ 새로운 엔티티: 걸스데이\\
\\
---\\
현재 엔티티: \{answer\}\\
\\
문장:\\
\{text\_to\_tts\}\\
\\
아래 형식으로만 출력하세요.\\
새로운 엔티티:
}
\end{tcolorbox}
\caption{Prompt used for stage 2 of the SCA-QA dataset construction pipeline. Given the original answer entity, the model generates a new entity of the same category that does not appear in the original context.}
\label{fig:sca_qa_stage2_prompt}
\end{figure*}

\begin{figure*}[t!]
\centering
\begin{tcolorbox}[
  width=\textwidth,
  title={\textbf{Stage 2: Entity Replacement --- English translation}\\\textit{\small Objective: Find a different entity of the same type to replace the original answer. Shown for readability; the Korean prompt in Figure~\ref{fig:sca_qa_stage2_prompt} is the one used.}},
  colback=gray!5,
  colframe=tealframe,
  colbacktitle=tealtitle,
  coltitle=black,
  fonttitle=\large,
  boxrule=0.5pt,
  top=2pt, bottom=2pt, left=4pt, right=4pt,
]

{\footnotesize\itshape
You are given a context and a current entity. Choose exactly one \textbf{different} entity of the \textbf{same type} as this entity and report it. (e.g., a historical battle $\rightarrow$ another battle name; a girl group $\rightarrow$ another girl group name.) Do not modify the sentence; output only the name of the new entity.\\
IMPORTANT!!! Do not use a word that appears in the context as the new entity. The new entity must not occur in the sentence.\\
\\
{[}Example 1{]}\\
Sentence: The Battle of Salsu was the battle that sealed the Second Goguryeo--Sui War as a Goguryeo victory. The 300,000-strong Sui army that had struck directly at Pyongyang was attacked by Eulji Mundeok's forces while crossing the Salsu River on its retreat, suffering devastating losses and fleeing in defeat. Together with the Battle of Gwiju and the Battle of Hansando, it is counted among the three great victories of Korean history.\\
Current entity: the Battle of Salsu\\
$\rightarrow$ New entity: the Battle of Ansi Fortress\\
Not allowed: the Battle of Hansando (it occurs in the context)\\
\\
{[}Example 2{]}\\
Sentence: Crayon Pop is a five-member South Korean girl group...\\
Current entity: Crayon Pop\\
$\rightarrow$ New entity: Girl's Day\\
\\
---\\
Current entity: \{answer\}\\
\\
Sentence:\\
\{text\_to\_tts\}\\
\\
Output in the following format only.\\
New entity:
}
\end{tcolorbox}
\caption{English translation of the prompt shown in Figure~\ref{fig:sca_qa_stage2_prompt}. The translation is provided for readability only; the Korean version is the one used in our experiments.}
\label{fig:sca_qa_stage2_prompt_en}
\end{figure*}

%% Stage 3 %%
\begin{figure*}[t!]
\centering
\begin{tcolorbox}[
  width=\textwidth,
  title={\textbf{Stage 3: Sentence Transformation}\\\textit{\small Objective: Replace the original entity with the new entity in the context.}},
  colback=gray!5,
  colframe=tealframe,
  colbacktitle=tealtitle,
  coltitle=black,
  fonttitle=\large,
  boxrule=0.5pt,
  top=2pt, bottom=2pt, left=4pt, right=4pt,
]
{\footnotesize\ttfamily
문장에서 `현재 엔티티'를 `새 엔티티'로 **바꾼 문장**을 만들어 주세요.\\
\\
{[}예시 1{]}\\
문장: 살수 대첩은 제2차 고수 전쟁을 고구려의 승리로 장식한 전투이다. 평양을 직공했던 수나라의 30만 군대가 살수를 건너 회군하던 중 을지문덕 군의 공격을 받아 궤멸적인 피해를 입고 패주했다. 귀주 대첩, 한산도 대첩과 함께 한국사 3대 대첩 중 하나로 불린다.\\
현재 엔티티: 살수 대첩\\
새 엔티티: 안시성 전투\\
→ 수정된 문장: 안시성 전투는 제2차 고수 전쟁을 고구려의 승리로 장식한 전투이다. 평양을 직공했던 수나라의 30만 군대가 살수를 건너 회군하던 중 을지문덕 군의 공격을 받아 궤멸적인 피해를 입고 패주했다. 귀주 대첩, 한산도 대첩과 함께 한국사 3대 대첩 중 하나로 불린다.\\
\\
{[}예시 2{]}\\
문장: 크레용팝은 대한민국의 오인조 걸 그룹이다. `크레용팝'이라는 이름은 음악이라는 도화지에 멤버들이 갖고 있는 다양한 색상을 입힌다는 의미를 담고 있다.\\
현재 엔티티: 크레용팝\\
새 엔티티: 걸스데이\\
→ 수정된 문장: 걸스데이는 대한민국의 오인조 걸 그룹이다. `걸스데이'라는 이름은 음악이라는 도화지에 멤버들이 갖고 있는 다양한 색상을 입힌다는 의미를 담고 있다.\\
\\
---\\
현재 엔티티: \{answer\}\\
새 엔티티: \{new\_entity\}\\
\\
문장:\\
\{text\_to\_tts\}\\
\\
아래 형식으로만 출력하세요.\\
수정된 문장:
}
\end{tcolorbox}
\caption{Prompt used for stage 3 of the SCA-QA dataset construction pipeline. The original answer entity is replaced with the newly generated entity to construct a counterfactual context.}
\label{fig:sca_qa_stage3_prompt}
\end{figure*}

\begin{figure*}[t!]
\centering
\begin{tcolorbox}[
  width=\textwidth,
  title={\textbf{Stage 3: Sentence Transformation --- English translation}\\\textit{\small Objective: Replace the original entity with the new entity in the context. Shown for readability; the Korean prompt in Figure~\ref{fig:sca_qa_stage3_prompt} is the one used.}},
  colback=gray!5,
  colframe=tealframe,
  colbacktitle=tealtitle,
  coltitle=black,
  fonttitle=\large,
  boxrule=0.5pt,
  top=2pt, bottom=2pt, left=4pt, right=4pt,
]

{\footnotesize\itshape
Produce a \textbf{modified sentence} in which the `current entity' in the sentence is replaced by the `new entity'.\\
\\
{[}Example 1{]}\\
Sentence: The Battle of Salsu was the battle that sealed the Second Goguryeo--Sui War as a Goguryeo victory. The 300,000-strong Sui army that had struck directly at Pyongyang was attacked by Eulji Mundeok's forces while crossing the Salsu River on its retreat, suffering devastating losses and fleeing in defeat. Together with the Battle of Gwiju and the Battle of Hansando, it is counted among the three great victories of Korean history.\\
Current entity: the Battle of Salsu\\
New entity: the Battle of Ansi Fortress\\
$\rightarrow$ Modified sentence: The Battle of Ansi Fortress was the battle that sealed the Second Goguryeo--Sui War as a Goguryeo victory. The 300,000-strong Sui army that had struck directly at Pyongyang was attacked by Eulji Mundeok's forces while crossing the Salsu River on its retreat, suffering devastating losses and fleeing in defeat. Together with the Battle of Gwiju and the Battle of Hansando, it is counted among the three great victories of Korean history.\\
\\
{[}Example 2{]}\\
Sentence: Crayon Pop is a five-member South Korean girl group. The name ``Crayon Pop'' carries the meaning of coloring the drawing paper of music with the various colors the members possess.\\
Current entity: Crayon Pop\\
New entity: Girl's Day\\
$\rightarrow$ Modified sentence: Girl's Day is a five-member South Korean girl group. The name ``Girl's Day'' carries the meaning of coloring the drawing paper of music with the various colors the members possess.\\
\\
---\\
Current entity: \{answer\}\\
New entity: \{new\_entity\}\\
\\
Sentence:\\
\{text\_to\_tts\}\\
\\
Output in the following format only.\\
Modified sentence:
}
\end{tcolorbox}
\caption{English translation of the prompt shown in Figure~\ref{fig:sca_qa_stage3_prompt}. The translation is provided for readability only; the Korean version is the one used in our experiments.}
\label{fig:sca_qa_stage3_prompt_en}
\end{figure*}

%% Stage 4 %%
\begin{figure*}[t!]
\centering
\begin{tcolorbox}[
  width=\textwidth,
  title={\textbf{Stage 4: TTS Text Normalization}\\\textit{\small Objective: Convert numerals and non-Korean characters into spoken Korean for TTS synthesis.}},
  colback=gray!5,
  colframe=tealframe,
  colbacktitle=tealtitle,
  coltitle=black,
  fonttitle=\large,
  boxrule=0.5pt,
  top=2pt, bottom=2pt, left=4pt, right=4pt,
]
{\footnotesize\ttfamily
아래 문장을 한국어 전사로 바꿔 주세요. 결과 문장은 한글만 포함해야 합니다.\\
{[}예시{]}\\
context:\\
황산벌 전투는 660년 8월 20일에서 8월 21일 황산벌에서 신라 군과 백제 군 사이에 일어났던 전투이다.\\
변환된 문장 : 황산벌 전투는 육백육십년 팔월 이십일에서 팔월 이십일일 황산벌에서 신라 군과 백제 군 사이에 일어났던 전투이다.\\
\\
아래 문장을 한국어 전사로 바꿔 주세요. 결과 문장은 한글만 포함해야 합니다.\\
context:\\
\{content\}\\
\\
변환된 문장:
}
\end{tcolorbox}
\caption{Prompt used for stage 4 of the SCA-QA dataset construction pipeline. Numerals and non-Korean expressions in the transformed context are normalized into spoken Korean forms for TTS synthesis.}
\label{fig:sca_qa_stage4_prompt}
\end{figure*}

\begin{figure*}[t!]
\centering
\begin{tcolorbox}[
  width=\textwidth,
  title={\textbf{Stage 4: TTS Text Normalization --- English translation}\\\textit{\small Objective: Convert numerals and non-Korean characters into spoken Korean for TTS synthesis. Shown for readability; the Korean prompt in Figure~\ref{fig:sca_qa_stage4_prompt} is the one used.}},
  colback=gray!5,
  colframe=tealframe,
  colbacktitle=tealtitle,
  coltitle=black,
  fonttitle=\large,
  boxrule=0.5pt,
  top=2pt, bottom=2pt, left=4pt, right=4pt,
]

{\footnotesize\itshape
Convert the sentence below into a Korean spoken-form transcription. The resulting sentence must contain Hangul characters only.\\
{[}Example{]}\\
context:\\
The Battle of Hwangsanbeol was fought between the Silla and Baekje armies at Hwangsanbeol from 20 to 21 August 660. \textnormal{(The Korean source writes the year and dates in Arabic numerals.)}\\
Converted sentence: the same sentence with every numeral spelled out in Korean words (e.g., ``660'' $\rightarrow$ ``yukbaengnyuksimnyeon'', ``August 20'' $\rightarrow$ ``paruol isibil'').\\
\\
Convert the sentence below into a Korean spoken-form transcription. The resulting sentence must contain Hangul characters only.\\
context:\\
\{content\}\\
\\
Converted sentence:
}
\end{tcolorbox}
\caption{English translation of the prompt shown in Figure~\ref{fig:sca_qa_stage4_prompt}. The translation is provided for readability only; the Korean version is the one used in our experiments.}
\label{fig:sca_qa_stage4_prompt_en}
\end{figure*}

%% Stage 5 %%
\begin{figure*}[t!]
\centering
\begin{tcolorbox}[
  width=\textwidth,
  title={\textbf{Stage 5: Distractor Generation}\\\textit{\small Objective: Generate two additional distractors to form four-way multiple-choice options.}},
  colback=gray!5,
  colframe=tealframe,
  colbacktitle=tealtitle,
  coltitle=black,
  fonttitle=\large,
  boxrule=0.5pt,
  top=2pt, bottom=2pt, left=4pt, right=4pt,
]
{\footnotesize\ttfamily
4지선다 문제를 만들기 위해 선택지 4개가 필요합니다. 정답은 이미 정해져 있습니다.\\
\\
정답 (new\_answer): \{new\_answer\}\\
원래 답 (answer, 오답으로 사용): \{answer\}\\
\\
위와 **같은 유형**의 비슷한 오답 2개를 더 만들어 주세요. (정답/원래 답과 같은 카테고리이지만 다른 항목)\\
예: 전투명 → 다른 전투명, 걸그룹명 → 다른 걸그룹명, 기관명 → 다른 기관명, 야구 선수 → 다른 야구 선수\\
\\
아래 형식으로만 출력하세요.\\
오답1: (한 단어 또는 짧은 구)\\
오답2: (한 단어 또는 짧은 구)
}
\end{tcolorbox}
\caption{Prompt used for stage 5 of the SCA-QA dataset construction pipeline. Two additional distractor entities from the same category are generated and combined with the original and replaced answers to construct four-way multiple-choice options.}
\label{fig:sca_qa_stage5_prompt}
\end{figure*}

\begin{figure*}[t!]
\centering
\begin{tcolorbox}[
  width=\textwidth,
  title={\textbf{Stage 5: Distractor Generation --- English translation}\\\textit{\small Objective: Generate two additional distractors to form four-way multiple-choice options. Shown for readability; the Korean prompt in Figure~\ref{fig:sca_qa_stage5_prompt} is the one used.}},
  colback=gray!5,
  colframe=tealframe,
  colbacktitle=tealtitle,
  coltitle=black,
  fonttitle=\large,
  boxrule=0.5pt,
  top=2pt, bottom=2pt, left=4pt, right=4pt,
]

{\footnotesize\itshape
Four options are needed to construct a four-way multiple-choice question. The correct answer has already been determined.\\
\\
Correct answer (new\_answer): \{new\_answer\}\\
Original answer (answer, used as a distractor): \{answer\}\\
\\
Generate two more similar distractors of the \textbf{same type} as the above (same category as the correct and original answers, but different items).\\
e.g., a battle name $\rightarrow$ another battle name; a girl group name $\rightarrow$ another girl group name; an institution name $\rightarrow$ another institution name; a baseball player $\rightarrow$ another baseball player.\\
\\
Output in the following format only.\\
Distractor 1: (a single word or short phrase)\\
Distractor 2: (a single word or short phrase)
}
\end{tcolorbox}
\caption{English translation of the prompt shown in Figure~\ref{fig:sca_qa_stage5_prompt}. The translation is provided for readability only; the Korean version is the one used in our experiments.}
\label{fig:sca_qa_stage5_prompt_en}
\end{figure*}

%% Stage 6 %%
\begin{figure*}[t!]
\centering
\begin{tcolorbox}[
  width=\textwidth,
  title={\textbf{Stage 6: Answer Verification}\\\textit{\small Objective: Verify that the generated question is answerable from the transformed context.}},
  colback=gray!5,
  colframe=tealframe,
  colbacktitle=tealtitle,
  coltitle=black,
  fonttitle=\large,
  boxrule=0.5pt,
  top=2pt, bottom=2pt, left=4pt, right=4pt,
]
{\footnotesize\ttfamily
아래 질문에 대한 답을 컨텍스트에서 찾아서 답하세요. 알고 있던 내용과 다르더라도 컨텍스트에 있는 내용을 기반으로 답하세요.\\
{[}예시{]}\\
컨텍스트: 안시성 전투는 제2차 고수 전쟁을 고구려의 승리로 장식한 전투이다. 평양을 직공했던 수나라의 30만 군대가 살수를 건너 회군하던 중 을지문덕 군의 공격을 받아 궤멸적인 피해를 입고 패주했다. 귀주 대첩, 한산도 대첩과 함께 한국사 3대 대첩 중 하나로 불린다.\\
질문: 제2차 고수 전쟁에서 고구려의 승리를 이끈 전투로, 수나라의 30만 군대가 을지문덕 군의 공격을 받아 큰 피해를 입고 패주한 전투는 무엇인가?\\
보기: (1) 백강 전투 (2) 살수 대첩 (3) 황산 전투 (4) 안시성 전투\\
답: (4) 안시성 전투\\
\\
\\
아래 질문에 대한 답을 컨텍스트에서 찾아서 답하세요. 알고 있던 내용과 다르더라도 컨텍스트에 있는 내용을 기반으로 답하세요.\\
컨텍스트: \{context\}\\
질문: \{question\}\\
보기: \{choices\}\\
답:
}
\end{tcolorbox}
\caption{Prompt used for stage 6 of the SCA-QA dataset construction pipeline. The generated sample is verified by checking whether the question can be correctly answered from the transformed context.}
\label{fig:sca_qa_stage6_prompt}
\end{figure*}

\begin{figure*}[t!]
\centering
\begin{tcolorbox}[
  width=\textwidth,
  title={\textbf{Stage 6: Answer Verification --- English translation}\\\textit{\small Objective: Verify that the generated question is answerable from the transformed context. Shown for readability; the Korean prompt in Figure~\ref{fig:sca_qa_stage6_prompt} is the one used.}},
  colback=gray!5,
  colframe=tealframe,
  colbacktitle=tealtitle,
  coltitle=black,
  fonttitle=\large,
  boxrule=0.5pt,
  top=2pt, bottom=2pt, left=4pt, right=4pt,
]

{\footnotesize\itshape
Find the answer to the question below in the context and answer accordingly. Even if it differs from what you already know, answer based on the content of the context.\\
{[}Example{]}\\
Context: The Battle of Ansi Fortress was the battle that sealed the Second Goguryeo--Sui War as a Goguryeo victory. The 300,000-strong Sui army that had struck directly at Pyongyang was attacked by Eulji Mundeok's forces while crossing the Salsu River on its retreat, suffering devastating losses and fleeing in defeat. Together with the Battle of Gwiju and the Battle of Hansando, it is counted among the three great victories of Korean history.\\
Question: Which battle led Goguryeo to victory in the Second Goguryeo--Sui War, in which the 300,000-strong Sui army was attacked by Eulji Mundeok's forces and fled after heavy losses?\\
Options: (1) the Battle of Baekgang (2) the Battle of Salsu (3) the Battle of Hwangsan (4) the Battle of Ansi Fortress\\
Answer: (4) the Battle of Ansi Fortress\\
\\
\\
Find the answer to the question below in the context and answer accordingly. Even if it differs from what you already know, answer based on the content of the context.\\
Context: \{context\}\\
Question: \{question\}\\
Options: \{choices\}\\
Answer:
}
\end{tcolorbox}
\caption{English translation of the prompt shown in Figure~\ref{fig:sca_qa_stage6_prompt}. The translation is provided for readability only; the Korean version is the one used in our experiments.}
\label{fig:sca_qa_stage6_prompt_en}
\end{figure*}

\subsection{Prompts for LLM as Judge.}
\label{sec:llm_as_judge}
We use an LLM as judge to automatically assess response quality for tasks that require open ended generation. 
Figure~\ref{fig:llm_judge_prompt} presents the prompt template used for SIF. 
The judge receives three fields: the input question $\{question\}$, the reference answer $\{reference\_answer\}$, and the model generated answer $\{generated\_answer\}$. 
It then compares the generated answer against the reference answer under three criteria: accuracy, completeness, and relevance. 
Accuracy measures whether the response correctly answers the question, completeness measures whether it provides sufficient information relative to the reference answer, and relevance measures whether the response remains aligned with the given question. 
To ensure a consistent and directly usable output, the judge is instructed to return only a scalar score between 0.0 and 1.0. An English translation of this prompt is provided in Figure~\ref{fig:llm_judge_prompt_en} for reference.

\subsection{Evaluation Prompts}
\label{sec:prompts_list}
We use task specific prompt templates for all KoALa-Bench evaluation tasks. 
For each task, we evaluate each model with four semantically equivalent prompt variants and report the best performing result under the four prompt reporting protocol. 
Figures~\ref{fig:eval_prompts_part1} and~\ref{fig:eval_prompts_part2} list the prompt variants used for general speech tasks and QA based tasks, respectively.
English translations of these prompts are provided in Figures~\ref{fig:eval_prompts_part1_en} and~\ref{fig:eval_prompts_part2_en} for reference.

\subsection{SCA-QA Dataset Construction Prompts}
\label{sec:sca_qa_prompts}
We provide the full prompts used in each stage of the SCA-QA construction pipeline (Algorithm~\ref{alg:sca_qa_pipeline}). 
Figures~\ref{fig:sca_qa_stage1_prompt}--\ref{fig:sca_qa_stage6_prompt} present the prompts for question and answer generation (Stage~1), entity replacement (Stage~2), sentence transformation (Stage~3), TTS text normalization (Stage~4), distractor generation (Stage~5), and answer verification (Stage~6), respectively.
English translations of all six prompts are provided in Figures~\ref{fig:sca_qa_stage1_prompt_en}--\ref{fig:sca_qa_stage6_prompt_en} for reference.

\end{CJK}
\end{document}